\ifdefined\pdfoutput\pdfoutput=1\fi
\documentclass[]{fairmeta}

\usepackage{amsmath,amssymb,amsthm}
\usepackage{graphicx}
\usepackage{booktabs}
\usepackage{multirow}
\usepackage{subcaption}
\usepackage{adjustbox}   % used by tables/
\usepackage{xspace}
\usepackage{float}       % [H] placement used throughout section/
\usepackage{wrapfig}     % wrapped (in-text) figures
\usepackage{tikz}        % native bar charts (ablation figure)

\usepackage[numbers,sort&compress]{natbib}      % author-year citations (Meta-report look)
\let\cite\citep           % plain \cite -> \citep: bracketed numeric [N] in numbers mode
\usepackage[colorlinks=true,breaklinks=true,
            linkcolor=metablue,citecolor=metablue,urlcolor=metablue]{hyperref}
\ifdefined\pdfsuppresswarningpagegroup\pdfsuppresswarningpagegroup=1\fi

\newcommand{\mname}{\ensuremath{\mathcal{N}_0}\text{-TWAM}\xspace}
\newcommand{\mlong}{Tactile World-Action Model\xspace}
\newcommand{\moT}{MoT\xspace} % Mixture-of-Transformers

\newcommand{\rmfont}{\fontfamily{RobotoMono-TLF}\selectfont} % RobotoMono for figure labels

\newtcolorbox{keybox}[1][]{colback=metabg,colframe=metablue,
  boxrule=0.6pt,arc=6pt,left=6pt,right=6pt,top=6pt,bottom=6pt,#1}

\definecolor{metablue}{HTML}{26527A}

\logo{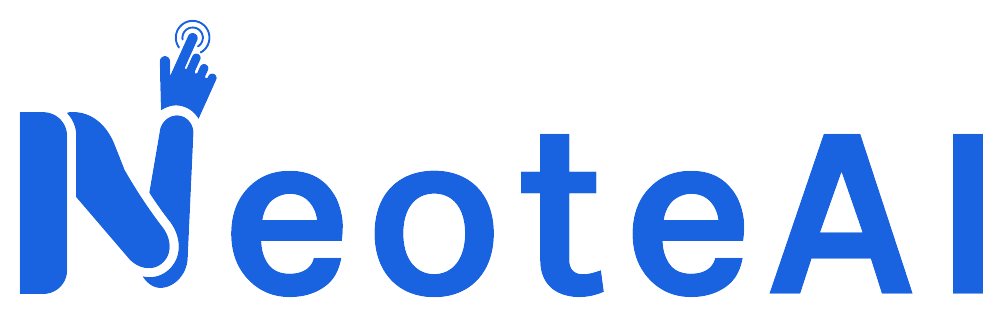}   % swap for logo-neoteai-cn.png / -mark.png; comment out for none
\logowidth{2.5cm}
\toplogo{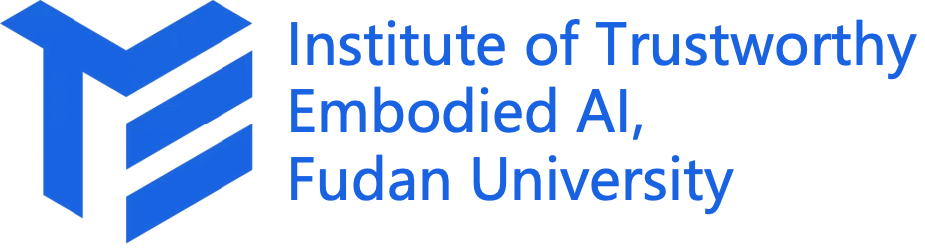}   % Fudan institute logo, stacked BELOW the neoteai logo
\toplogowidth{2.5cm}

\title{\mname{}: Scaling Tactile-Native World \\ Action Model for Contact-Rich Manipulation}

\author{\large NeoteAI Team \& Fudan TEAI Team}
\date{July 25, 2026}
\metadata[Code]{\url{https://github.com/neoteai/N0-TWAM}}
\metadata[Website]{\url{https://research.neoteai.com/n0-twam/}}

\begin{document}

\abstract{% ============================ ABSTRACT ============================
We present \mname{}, a tactile-native world-action model for contact-rich manipulation that predicts both future vision and contact. To our knowledge, it is the first tactile world-action model trained at large scale, and it demonstrates strong capability on contact-rich tasks. 
We pre-train \mname{} at large scale with visuo-tactile joint training over tactile-rich demonstrations, spanning six embodiments and $450$ tasks. 
We use NeoForce, a unified force-based tactile representation to form a physically grounded contact signal as condition for action generation. 
To boost long-horizon and multi-stage manipulation, we introduce tactile contact events for task staging and advance through them during execution.
For real-time efficiency, we adopt an asymmetric Mixture-of-Transformers architecture on \mname{}, which pairs a full-width expert for versatile videos and slim architecture for downstream action and tactile experts. Evaluations on both real and simulated benchmarks justify the powerful capabilities of \mname{} in various contact-rich tasks, as well as demonstrate the benefit of data scaling on precise tactile and action prediction. In summary, \mname{} endows a world-action model with predictive capabilities to foresee vision, tactile and action, building a solid foundation for fine-grained manipulation on open contact-rich tasks. The codebase and model checkpoints of \mname{} will be made publicly available to foster further research and development in tactile-enabled robotic manipulation.
}
\maketitle

% --- Body: one file per section under section/ ------------------------------
% ============================ 1. INTRODUCTION ============================
\section{Introduction}
\label{sec:intro}

Learning a general manipulation policy is hard in contact-rich settings. Seating a screw,
peeling one cup off a stack, or closing a gripper
on a soft object are decided less by the global layout of the scene than by what is
happening at the fingertips, a distinction that a camera can rarely resolve but that
a force or tactile sensor reports directly. A policy that is to act well in these
regimes therefore needs two things that pure visuomotor regression does not naturally
provide: access to touch, and the ability to \emph{anticipate} how the immediate
future of an interaction will unfold rather than reacting only to the present frame.

\begin{figure}[t]
    \centering
    \includegraphics[width=\textwidth]{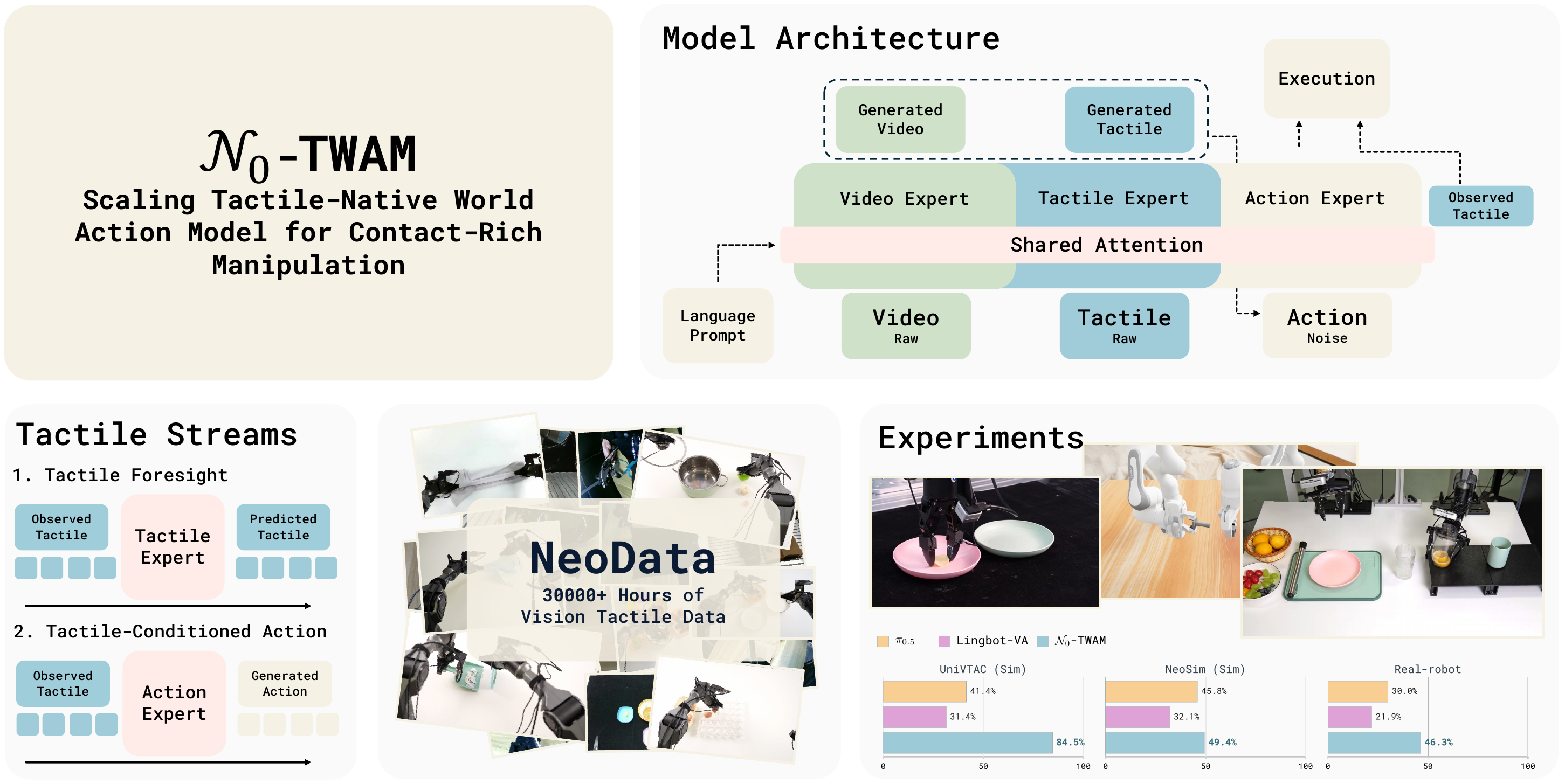}
    \caption{\textbf{Overview.} \mname{} is pre-trained on large-scale self-collected real-robot manipulation data with synchronized per-finger tactile
    streams. An MoT with three per-modality experts (video, tactile,
    and action) predicts the future vision and tactile, then predicts the action from
    them, so touch is both \emph{predicted} (a predicted future) and \emph{observed} (an
    observed present). Across contact-rich benchmarks in simulation and on real robots,
    \mname{} outperforms prior world-action and vision-language-action baselines.}
    \label{fig:front}
\end{figure}

Two recent lines of work each supply one of these ingredients. Vision-language-action
(VLA) models attach an action head to a large pretrained vision-language model and map
observations to actions, inheriting broad semantic priors but compressing a rich
visual present into a low-dimensional action with no explicit model of what happens
next~\cite{pi0}. Video world models, on the other hand, learn to roll the future of an
observation forward in time, and a growing body of \emph{world-action} models couples
such a video predictor with action so that control is grounded in a predicted
future~\cite{lingbotva,dreamzero}. The latter framing is attractive for contact-rich
control (if the model can predict the next moment, it can decide against it), yet
the predicted future of these models is almost always purely visual.

Touch is beginning to enter this picture, but so far along three partial paths.
Tactile policies \emph{consume} touch as an input channel
and never predict it~\cite{tactilevla}. A second family forecasts touch with a
separate (often frozen) tactile predictor bolted onto the policy, so prediction occurs happens outside the network that acts~\cite{dreamtacvla,tacforesight}. And
concurrent touch-aware world-action models admit touch into the model but keep it at
arm's length, gating or masking tactile tokens to protect the visual
stream~\cite{dreamtac,tactilewam,omnivta}. On every path, touch is not part of what the model predicts. What is missing is a model
that predicts touch together with vision, under the same objective and at the same causal
step, and a policy that reads its actions from that jointly predicted future.
\iffalse % Figure 2 (teaser) 暂时撤下，保留以便日后恢复
\begin{figure}[H]
    \centering
    \includegraphics[width=\textwidth]{Figure/teaser_paths.pdf}
    \caption{\textbf{Four ways to put touch into a policy.} (a)~Tactile policies
    consume touch as an input channel but never predict it~\cite{tactilevla}.
    (b)~A second family forecasts touch with a separate, often frozen, predictor
    bolted onto the policy~\cite{dreamtacvla,tacforesight}. (c)~Concurrent
    touch-aware world-action models admit touch into the model but keep it at
    arm's length: gated attention or separate streams shield the visual
    dynamics~\cite{dreamtac,tactilewam,omnivta}. (d)~\mname{} models touch jointly
    with vision: one network predicts future vision and touch
    jointly, under one objective, and reads the action out of that predicted
    future.}
    \label{fig:teaser}
\end{figure}
\fi

We close this gap with \mname{} (\mlong{}), a world-model policy and, to our knowledge, the first tactile world-action model trained at large scale. \mname{} is built on a video-diffusion transformer backbone and keeps the world-action recipe of jointly modeling future observations and action, but it predicts \emph{tactile as part of the predicted future} rather than treating it as a side input. The backbone is reorganized as a Mixture-of-Transformers (MoT)~\cite{mot}: three per-modality experts (video, action and tactile) each keeps their own weights and capacity and interact only through a single shared self-attention at every layer. Where concurrent touch-aware models protect their visual stream by restricting touch in \emph{attention}, \mname{} isolates capacity in \emph{weights}, so vision and touch stay fully attentive to one another. A frame-id causal schedule turns this shared attention into a ``predict-then-act'' cascade inside one forward pass: within a chunk the video and tactile experts co-generate the future scene and the future contact, and the action expert is conditioned on those just-predicted quantities. That future contact is the predicted tactile; as it acts, the action expert also reads the observed tactile, cross-attended in from the current reading. The policy therefore acts on touch it has already predicted, grounded in the touch it feels now.

Treating three modalities as full experts would, however, triple the cost of the
largest component, and the action and tactile streams have no large-scale pretraining
to justify that cost. \mname{} instead uses an \emph{asymmetric} design: the
video expert keeps its full, original width, while the action and tactile experts are slim
and trained from scratch, benefiting from
the visual prior only indirectly through the shared attention.  This halves the parameter count relative to an all-full-width design, and the same asymmetry keeps inference cheap: once a chunk's video and tactile are predicted, their attention keys and values are cached, so each action-denoising step re-runs only the lightweight action expert rather than the full-width video expert.

Across simulated and real contact-rich benchmarks, \mname{} is the strongest method
overall: it reaches $84.5\%$ on UniVTAC and $49.4\%$ on NeoSim in simulation, and averages
$46.3\%$ across our eight-task real-robot suite, against $30.0\%$ for the strongest
vision-language-action baseline and $21.9\%$ and $14.4\%$ for the vision-only world-action
baselines. Ablations attribute the gains to both tactile roles (removing either the predicted
foresight target or the observed conditioning degrades success), and under distribution shift
the policy degrades more gracefully than vision-only world-action baselines. We summarize our
contributions as follows.

% \paragraph{Contributions.}
\begin{itemize}
  \item \textbf{A native tactile world-action model (\S\ref{sec:arch}).} We present
        \mname{}, a world-action model with touch as a \emph{native} modality of its MoT backbone: a dedicated tactile expert predicts future
        touch jointly with future vision under one shared self-attention, rather than as
        a side input. We pre-train it at scale on tens of thousands hours of real-robot
        data (much of it carrying synchronized tactile), spanning six embodiments and
        $450$ tasks, collected with our own system.
  \item \textbf{A dual-pathway predictive tactile design (\S\ref{sec:tactile}).} We
        model touch inside the world-action model along two pathways: a
        \emph{predicted} tactile stream the model generates as a foresight target, and an
        \emph{observed} tactile stream that conditions action generation. In
        post-training, the observed pathway reads touch through \emph{NeoForce}~\cite{neodata},
        effectively extracting force representation on manipulation tasks.
   \item \textbf{A tactile-aware sub-task system for long-horizon manipulation (\S\ref{sec:inference}).} We use tactile contact events, such as a grasp or a release, to segment long-horizon demonstrations into sub-task-conditioned clips for training and to schedule execution stages at inference: the predicted tactile triggers the advance to the next sub-task, and the observed contact event confirms it, so both tactile roles drive the planner.
   \item \textbf{Strong empirical results (\S\ref{sec:exp}).} \mname{} achieves
        state-of-the-art results on contact-rich benchmarks in simulation and on real
        robots. Our analyses trace these gains to touch itself: both tactile pathways
        contribute, performance grows with more pre-training data, and it stays robust
        under distribution shift.
\end{itemize}

% ============================ 4. MODEL ============================
\section{Model}
\label{sec:model}

\mname{} is a world-action model: rather than regress actions from observations, it
learns one generative model that rolls the future the robot will see and feel forward in
time and reads the action out of that predicted future. Three choices set it apart from a
video world-action model. (i)~\emph{Touch is modeled jointly with vision}: the future tactile
signal is generated under the same flow-matching objective as the future video, at the
same causal step, so the model predicts touch directly rather than deriving it from pixels. (ii)~\emph{Capacity is isolated in weights, not attention}:
each modality gets its own expert, and the experts share a single self-attention, so vision and touch keep private weights yet stay
fully attentive to one another; concurrent touch-aware models instead gate or mask
tactile tokens to protect the visual stream. (iii)~\emph{Touch plays two roles}: it is
both predicted ahead of time, as a foresight target, and observed now, as an observation
the action reads. We set up the architecture (\S\ref{sec:arch}), develop the
two tactile roles (\S\ref{sec:tactile}), and close with tactile-aware execution
(\S\ref{sec:inference}).

\begin{figure}[t]
    \centering
    \includegraphics[width=\textwidth]{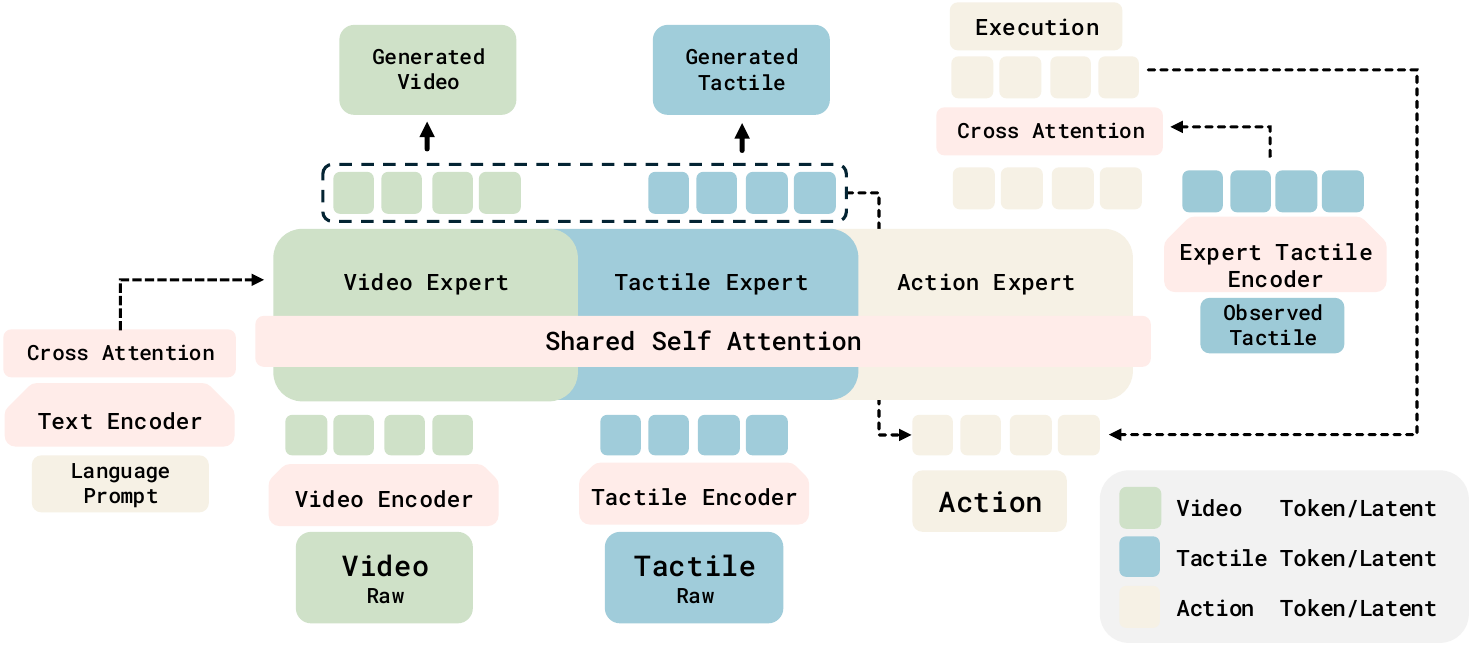}
    \caption{\textbf{Overview of \mname{}.} Multi-view RGB and tactile streams are encoded into latent tokens and processed by three per-modality experts (\emph{video}, \emph{tactile}, \emph{action}) that interact only through a single shared self-attention at every layer, ordered by the frame-id causal cascade: the video and tactile experts co-generate the predicted future scene and future contact, and the action expert denoises the action chunk conditioned on the just-predicted pair. The language instruction enters the video and
    action experts through cross-attention, but not the tactile expert. A lightweight observed pathway
    (right) encodes the \emph{observed} current tactile and cross-attends into the
    action tokens before the action head (\S\ref{sec:tactile}); the denoised actions
    are decoded and executed.}
    \label{fig:overview}
\end{figure}
% =========================================================================
\subsection{Model architecture}
\label{sec:arch}\label{sec:backbone}

\paragraph{Problem setup.}
Conditioned on a language instruction and the observation history, \mname{} predicts
three coupled streams of the robot's near future: what it will \emph{see}, what it will
\emph{feel}, and the \emph{action} it will take.
Given the observation history and the instruction $c$, \mname{} predicts each chunk's future
video $x^{v}_{k}$, tactile $x^{t}_{k}$, and action $x^{a}_{k}$, and models their joint distribution
autoregressively over chunks:
\begin{equation}
p_\theta\big(x^{v}_{1:K},\,x^{t}_{1:K},\,x^{a}_{1:K}\,\big|\,c\big)
\;=\;
\prod_{k=1}^{K}\;
\underbrace{p_\theta\big(x^{v}_{k},\,x^{t}_{k}\,\big|\,X_{<k},\,c\big)}_{\text{predict}}
\;\cdot\;
\underbrace{p_\theta\big(x^{a}_{k}\,\big|\,x^{v}_{k},\,x^{t}_{k},\,X_{<k},\,c\big)}_{\text{act}},
\label{eq:factor}
\end{equation}
where the observation history $X_{<k}=(x^{v}_{<k},x^{t}_{<k},x^{a}_{<k})$ collects the
past video, tactile, and actions. \mname{} treats vision and touch as equally important and predicts the two jointly. The noisy action token is denoised
conditioned on the just-predicted $x^{v}_{k}$ and $x^{t}_{k}$ of its own chunk, so the
policy acts on an anticipated future.

\paragraph{Conditional flow-matching objective.}
Flow matching~\cite{flowmatching} regresses the velocity of the straight path from a clean
target $x$ to Gaussian noise $\epsilon\sim\mathcal{N}(0,I)$. We apply it at
\emph{latent-frame} granularity. Let $x^m_{k,j}$ denote temporal group $j$ of modality
$m\in\{v,a,t\}$ in chunk $k$: a VAE latent frame for video and tactile, and the
action steps aligned with each latent frame for action, so $J_v=J_t=J_a=2$. For each
group we independently sample $\sigma^m_{k,j}\sim p(\sigma)$ and
$\epsilon^m_{k,j}\sim\mathcal{N}(0,I)$:
\begin{equation}
\hat{x}^{m}_{k,j}
\;=\;
(1-\sigma^{m}_{k,j})\,x^{m}_{k,j}
\;+\;
\sigma^{m}_{k,j}\,\epsilon^{m}_{k,j},
\qquad
u^{m}_{k,j}
\;=\;
\epsilon^{m}_{k,j}-x^{m}_{k,j}.
\label{eq:interp}
\end{equation}
Within a group, the scalar $\sigma^m_{k,j}$ is shared by all its tokens (the spatial
tokens of a video or tactile latent frame, or the aligned action steps) while the
entries of $\epsilon^m_{k,j}$ remain independent. Writing
$\hat{X}=\{\hat{x}^m_{k,j}\}$ and $\Sigma=\{\sigma^m_{k,j}\}$, the loss is
\begin{equation}
\mathcal{L}
\;=\;
\sum_{m\in\{v,a,t\}}
\frac{\lambda_m}{J_m}
\sum_{j=1}^{J_m}
\mathbb{E}
\Big[
\big\|
\sqrt{w^m_{k,j}}\odot
\big(f^m_{\theta,k,j}(\hat{X},\Sigma,c;\mathcal{M})-u^m_{k,j}\big)
\big\|_2^2
\Big],
\label{eq:loss}
\end{equation}
where $w^m_{k,j}\ge0$ combines SNR weighting with the action-validity mask that drops the
absent arm on single-arm embodiments.

\paragraph{Mixture-of-Transformers backbone.}
Following the MoT design~\cite{mot}, we split the backbone into three
\emph{experts}, one per modality, that share \emph{only} a single self-attention at each layer (Fig.~\ref{fig:overview}). At layer
$\ell$, expert $m$ normalizes its own hidden state, applies its own timestep modulation,
and projects into a common attention width $d$:
\begin{equation}
Q^{m},K^{m},V^{m} \;=\; \mathrm{Proj}^{m}_{\ell}\!\big(\mathrm{AdaLN}^{m}_{\ell}(h^{m})\big),
\qquad m\in\{v,a,t\}.
\end{equation}
The per-expert queries/keys/values are concatenated in the fixed order $[\,v\,|\,a\,|\,t\,]$
and run through one masked self-attention,
\begin{equation}
[\,O^{v}\,|\,O^{a}\,|\,O^{t}\,] \;=\;
\mathrm{Attn}\!\big([\,Q^{v}|Q^{a}|Q^{t}],\,[\,K^{v}|K^{a}|K^{t}],\,[\,V^{v}|V^{a}|V^{t}];\,\mathcal{M}\big),
\end{equation}
after which each expert applies its own gated output projection and feed-forward network.

\paragraph{Parameter-efficient experts.}
A naive three-expert backbone would triple the cost of its largest component. We instead
keep only the shared attention at full width $d$ and let each expert run a narrower private
residual/FFN width $d_m$: thin boundary projections leave the concatenated attention
unchanged while the FFN, norms, and residual stream run at $d_m$. The full-width
\emph{video} expert is warm-started from a pretrained non-MoT, shared-backbone world-action model~\cite{lingbotva} reorganized here into experts, so the backbone inherits
a strong visual and dynamics prior. The slim \emph{action} and \emph{tactile} experts run
at $d_a{=}d_t{=}1024$ against the video expert's $d_v{=}3072$; not matching the shared
width, they are trained from scratch and reach the visual prior only through the shared
attention. Because the video expert alone is about $70\%$ of the backbone, slimming the two
new modalities rather than widening them roughly halves the trainable model
(Table~\ref{tab:params}).

\begin{table}[h]
\centering
\caption{Trainable \moT{} backbone configuration. The asymmetric design keeps a full-width
video expert (warm-started from a pretrained video model) alongside slim action and tactile
experts (trained from scratch), roughly halving the parameters of an all-full-width variant.
Shared self-attention runs at $d{=}3072$ ($24{\times}128$ heads) across all layers;
``Hidden Size'' is each expert's private residual/FFN width $d_m$. The frozen Wan VAE and
umT5 text encoder sit outside the backbone and are excluded; all-full-width figures are
estimates.}
\label{tab:params}
\small
\setlength{\tabcolsep}{4pt}
\begin{tabular}{lc cc cc cc c}
\toprule
 & & \multicolumn{2}{c}{Video} & \multicolumn{2}{c}{Action} & \multicolumn{2}{c}{Tactile} & \\
\cmidrule(lr){3-4}\cmidrule(lr){5-6}\cmidrule(lr){7-8}
Model & \shortstack{\#\\Layers} & \shortstack{Hidden\\Size} & Params & \shortstack{Hidden\\Size} & Params & \shortstack{Hidden\\Size} & Params & \shortstack{Total\\Params} \\
\midrule
\mname{} (asym.) & $30$ & $3072$ & $5.00$\,B & $1024$ & $1.13$\,B & $1024$ & $1.03$\,B & $7.16$\,B \\
All-full-width   & $30$ & $3072$ & $5.00$\,B & $3072$ & ${\approx}5.0$\,B & $3072$ & ${\approx}5.0$\,B & ${\approx}15$\,B \\
\bottomrule
\end{tabular}
\end{table}

\paragraph{Diffusion-forcing cascade.}
\begin{wrapfigure}{r}{0.46\textwidth}
    \centering
    \vspace*{-10pt}
    \begin{minipage}[c]{0.335\textwidth}
        \centering
        \includegraphics[width=\linewidth]{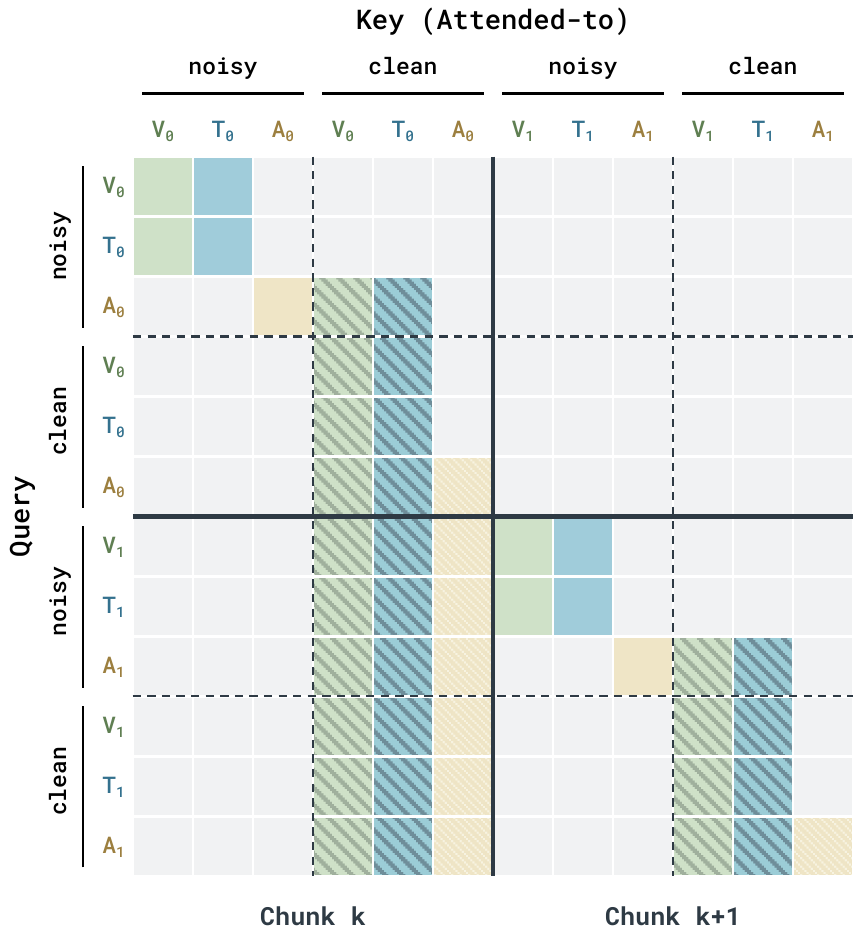}
    \end{minipage}\hfill
    \begin{minipage}[c]{0.105\textwidth}
        \centering
        \includegraphics[width=\linewidth]{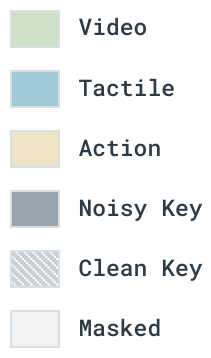}
    \end{minipage}
    \caption{\textbf{Joint self-attention mask} over two chunks, with each V/T block
    schematically collapsing the two latent frames. Each token appears as a
    key twice, noisy (solid) and clean (hatched); rows are queries, columns keys, colored
    by the key's modality. Within a chunk the noisy video and tactile co-generate with each
    other, and the noisy action additionally attends to their clean copies; each new chunk
    attends to the whole clean history, and no token ever attends a future chunk.}
    \label{fig:attn_mask}
\end{wrapfigure}
The shared attention is governed by a causal mask $\mathcal{M}$ that realizes the
factorization of Eq.~(\ref{eq:factor}) inside one forward pass. Two indices order every
token: a latent-frame position and a noise state. Within a chunk, its two latent frames
take consecutive positions; vision and touch tokens from the same frame share one
position, and the action follows the predicted frames on which it conditions. In state,
each token inherits the noise level of its latent-frame group and is \emph{noisy}
while it is being generated and \emph{clean} once it has become history. At training time
these clean history tokens are the ground-truth past: the model is \emph{teacher-forced} on
the true history and only denoises the current chunk, following the diffusion-forcing
regime~\cite{diffusionforcing} in which independent temporal-group noise levels generalize
teacher forcing to continuous tokens. As a robustness measure, the clean video and tactile
history is lightly re-noised with probability $0.5$, so the model learns to tolerate a
committed history that at test time is imperfect rather than ground truth; the action
history stays exactly clean. The mask then follows three rules: (i)~a clean token
attends to clean tokens at or before its position; (ii)~a noisy token attends to the clean
history strictly before its position; and (iii)~a noisy token attends to noisy tokens at
its \emph{own} position, i.e.\ co-generation.

Figure~\ref{fig:attn_mask} visualizes this mask for two chunks. The predict-then-act
cascade of Eq.~(\ref{eq:factor}) is thus produced entirely by the mask, with no change to
the attention operator itself. Because every temporal group draws its noise level
independently, training covers all combinations of clean and noisy groups; the two
patterns used at inference (a clean committed history with a noisy current chunk, and
clean predicted frames with a still-noisy action) are simply two of these combinations.

\paragraph{Action representation.}
Each arm contributes a $10$-dimensional per-frame action (a $3$-D end-effector position, a
$6$-D orientation~\cite{zhou2019continuity}, and a $1$-D gripper command), so a bimanual
action is $20$-dimensional; single-arm embodiments fill one half and the rest is masked
out of the loss. Following $\pi_{0.5}$~\cite{pi05}, we regress \emph{delta} end-effector
targets rather than absolute poses, anchored to the chunk-start pose, forming the delta by
plain element-wise subtraction: $\Delta p = p_{\text{t}}-p_{\text{a}}$ and
$\Delta r = r_{\text{t}}-r_{\text{a}}$, a coordinate-wise difference rather than a relative
rotation $R_{\text{a}}^{\top}R_{\text{t}}$ on $\mathrm{SO}(3)$. We prefer subtraction
because a relative rotation is a nonzero constant when the pose is nearly stationary, so a
small prediction error accumulates into a persistent orientation offset, whereas
subtraction makes the rest state the zero vector $\Delta r{=}0$, the natural target of a
regressor. This keeps the targets small and centered wherever the arm is; the gripper command stays absolute, and this $20$-D space is the
unified action interface used for pre-training across all embodiments.

% =========================================================================
\subsection{Modeling touch}
\label{sec:tactile}

The tactile sensors we use are \emph{vision-based}: a tactile sensor is essentially a small
camera imaging a soft gel that deforms on contact, so a tactile observation is itself an RGB
video stream. \mname{} therefore treats scene video and touch as equally important
modalities: the same VAE encodes them and the same generative objective predicts them.

Contact states such as a secure grasp, incipient slip, or excessive force are decisive
for the next action yet often invisible to the external scene cameras. \mname{} therefore
gives touch two roles: a future the model \emph{predicts} (\S\ref{sec:tactile-foresight})
and a present the policy \emph{reads} as it acts (\S\ref{sec:tactile-local}).

\subsubsection{Predicting future touch}
\label{sec:tactile-foresight}
Each tactile stream is VAE-encoded~\cite{wan2024} and patchified with the same $(1,2,2)$ patch as the video latent,
so touch enters the backbone as ordinary latent tokens;
a learned \emph{sensor-id} embedding (up to four sensors) lets one expert serve single-arm and bimanual data unchanged.
Sharing the latent space with video is what makes the shared attention meaningful, since vision and touch tokens become directly comparable.
Touch nonetheless keeps \emph{private weights}: tactile statistics are sparse and event-driven,
and concurrent touch-aware world models report that injecting such tokens into a visual dynamics model degrades video and action prediction~\cite{tactilewam}.
Isolating touch at the \emph{weight} level, rather than restricting it in \emph{attention}, lets vision and touch stay mutually attentive while keeping separate capacity.
The tactile expert also does not cross-attend to text.

The tactile expert treats the \emph{predicted} tactile latent as a generation target alongside video:
its two latent frames are independently noised and optimized by Eq.~(\ref{eq:loss});
this makes the model predict future contact a chunk ahead rather than only reading the current tactile signal.
We predict future touch in \emph{residual} form. Let $x^{t}_{i}$ be the tactile latent at
future step $i$ and $x^{t}_{0}$ the initial (step-$0$) frame; the tactile expert predicts the
residual
\begin{equation}
\Delta^{t}_{i} \;=\; x^{t}_{i} - x^{t}_{0},
\qquad\text{so that}\qquad
x^{t}_{i} \;=\; x^{t}_{0} + \Delta^{t}_{i},
\label{eq:tactile-residual}
\end{equation}
rather than the absolute frame, since touch is near-constant away from contact and its
informative content is the change $\Delta^{t}_{i}$ at contact onset and release.
Each tactile token takes the temporal position of the video frame it
accompanies, so it obeys the same causal and windowing rules with no tactile-specific
bypass. Two consequences follow. First, the predicted scene and predicted contact are denoised
\emph{together} under rule~(iii) of \S\ref{sec:arch}, each conditioning the other, so the
predicted video and predicted contact stay consistent. Second, a noisy tactile token reads
only committed history, never a same-step clean copy of itself, so no separate leak-guard
rule is needed once touch is aligned to the same frames as the video.

\subsubsection{Conditioning on observed touch}
\label{sec:tactile-local}
\begin{figure}[H]
    \centering
    \includegraphics[width=\textwidth]{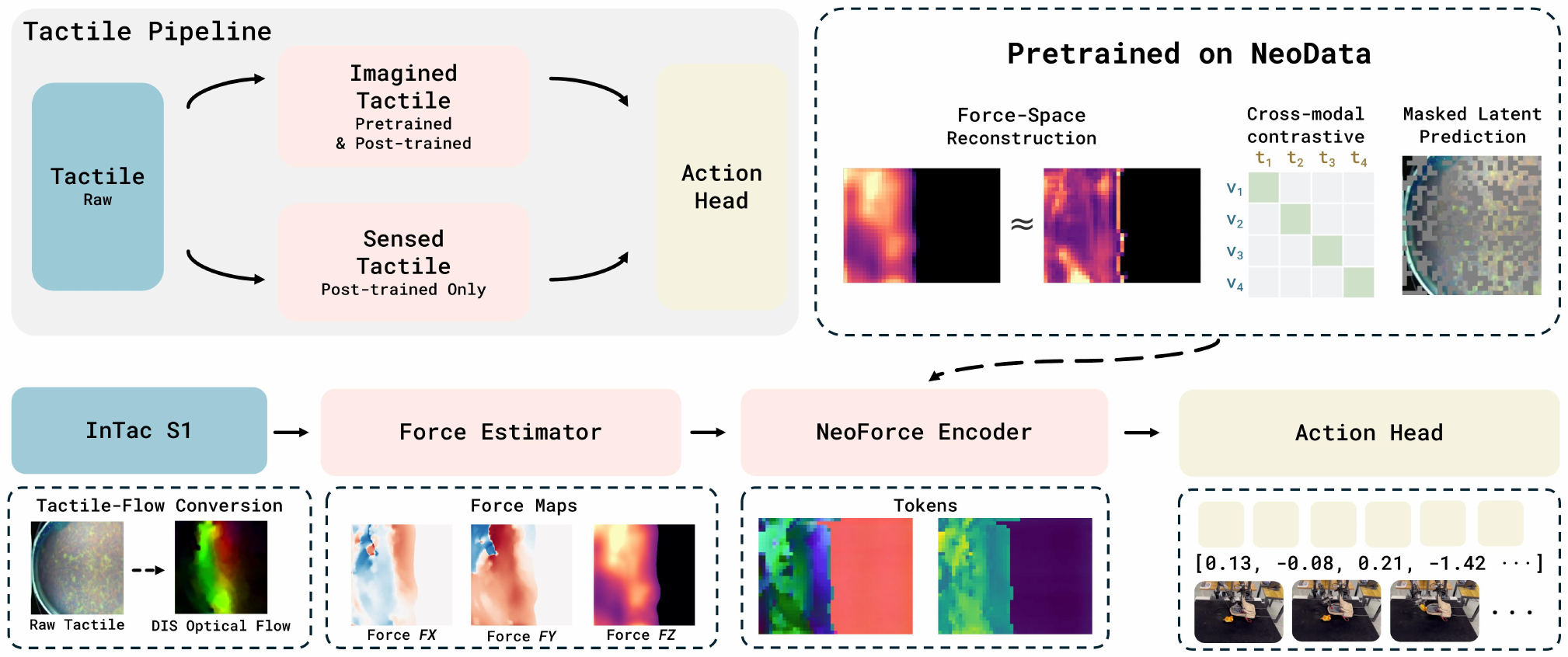}
    \caption{\textbf{The two tactile pathways.} Touch enters \mname{} in two ways. As a
    \emph{predicted} signal, it is generated in the VAE latent space shared with video and
    denoised by the tactile expert as a foresight target. As an \emph{observed} signal, it
    is read in force
    space: a frozen estimator $g_\theta$ maps the raw tactile image to a three-axis
    force map $[f_x,f_y,f_z]$ $\to$ a NeoForce~\cite{neodata} tactile encoder
    (warm-started, fine-tuned) produces representation tokens $\to$ cross-attended
    into the action stream through the zero-initialized port of
    Eq.~(\ref{eq:local}). An optional reconstruction term anchors the fine-tuned
    representation to the force field.}
    \label{fig:two_spaces}
\end{figure}

Touch reaches the action expert in \emph{two} representations, each matched to its role
(Fig.~\ref{fig:two_spaces}). The \emph{predicted} pathway is anticipatory: the predicted
future contact flows to action through the cascade, and because it must be
\emph{generated} it lives in the VAE latent space shared with video. The \emph{observed}
pathway is reactive, and the action head need only \emph{understand} the current reading,
not generate it, so we read it in \emph{force space}. A frozen estimator maps the raw
tactile image to a dense three-axis surface force map $[f_x,f_y,f_z]$ with a contact
mask, a physically grounded, sensor-agnostic reading of contact; a NeoForce tactile
encoder~\cite{neodata}, fine-tuned end-to-end with the action loss,
turns this map into representation tokens. Those tokens are cross-attended into the action
stream just before the action head, under a chunk-causal mask:
\begin{equation}
h^{a} \;\leftarrow\; h^{a} \;+\;
\mathrm{CrossAttn}\big(h^{a},\; E_{\mathrm{obs}}(x^{t}_{\mathrm{now}})\big),
\qquad W_{\mathrm{out}}=0 \;\text{at initialization},
\label{eq:local}
\end{equation}
so the pathway starts as an exact no-op and grafts onto a pretrained model without
perturbing it. In short, \mname{} \emph{predicts} touch in latent space and \emph{observes}
it in force space. The observed-tactile encoder $E_{\mathrm{obs}}$ is representation-agnostic, and we
instantiate it per domain. On real robots the tactile sensor matches the one NeoForce was
pretrained on, so $E_{\mathrm{obs}}$ is the NeoForce force-space encoder. In
simulation the tactile comes from a different sensor that NeoForce never saw, so we do not
read it in force space; instead $E_{\mathrm{obs}}$ is a lightweight branch trained from
scratch on the current-frame tactile latent: a linear patch embedding with learned
sensor-id and frame/height/width position embeddings whose tokens enter the same
zero-initialized cross-attention port, with no force estimator and no pretrained
representation.

\paragraph{Staged training.}
We separate the two roles across the two training stages (\S\ref{sec:impl}).
Pre-training uses \emph{only} the predicted stream, with the observed pathway off: its job is
to learn a generative world model at scale, and the predicted tactile latent is a generation
target that fits that objective, forcing the model to predict future contact jointly with
the future scene and so learn contact dynamics across the large-scale corpus. The observed stream
does not fit that objective. It is a conditioning input rather than a generation target,
it depends on the pretrained NeoForce encoder and on paired current-tactile readings, and,
most importantly, giving the policy the true current contact during pre-training would
let it rely on that signal and under-train the foresight it is meant to complement.
Post-training then switches on the observed pathway of Eq.~(\ref{eq:local}), adding it on
top of a model that has already learned to predict future contact; the ablations of
\S\ref{sec:ablations} separate the two contributions.

\paragraph{The pretrained force-space representation (NeoForce).}
The encoder of the observed pathway is \emph{NeoForce}, the tactile
representation model of $\mathcal{N}_0$-Foundation~\cite{neodata}, summarized here only
as far as \mname{} needs it.

NeoForce takes a synchronized chunk of RGB observations and tactile force maps and returns a
tactile representation in \emph{force space}. Its tactile input is not a raw gel image
but a dense three-axis surface force map
$\hat{F}=g_\theta(T)\in\mathbb{R}^{H\times W\times 3}$ produced by a learned estimator
$g_\theta$. The in-plane components $[f_x,f_y]$ carry the shear that drives sliding and
grasping, the normal component $f_z$ carries the pressure of pressing, and a binary
contact mask marks where the surface is touched, giving six channels for a parallel
gripper. Because these units are physical rather than tied to the appearance of a
particular gel, one encoder can serve different sensors~\cite{ftp1}. The visual and tactile streams
are patchified independently and fused by a shared ViT backbone initialized from DINOv2
weights, so vision and touch are read in a common space and the representation captures
how contact evolves across the chunk. A reconstruction head decodes the force map and
contact mask back out, which keeps the representation tied to a physical quantity.

\mname{} imports both stages and treats them differently. The estimator $g_\theta$ is a
calibrated sensor-to-physics converter and stays \emph{frozen}, so gradients stop at the
force map. The NeoForce encoder warm-starts $E_{\mathrm{obs}}$ in Eq.~(\ref{eq:local}) and
is \emph{fine-tuned} with the action loss, which lets the policy specialize contact
priors learned at scale to the behaviors it is learning. Its reconstruction head comes
along to supply the optional force-space anchor noted above.

% =========================================================================
\subsection{Tactile-aware execution}
\label{sec:inference}

At test time \mname{} runs autoregressively over chunks: within each chunk it follows
the cascade of Fig.~\ref{fig:attn_mask}, denoising the future video and tactile latents,
then the action conditioned on those just-predicted quantities, then decoding and executing
before sliding on. The autoregressive rollout, its rolling key/value cache, and the
asynchronous prediction pipeline are standard machinery for video
world-action models. What is specific to \mname{} is that inference is
\emph{tactile-aware}: touch enters the closed loop on three time scales at once, a reflex
at sensor rate (the observed pathway of Eq.~(\ref{eq:local})), a prediction at chunk rate
(the tactile foresight denoised alongside video), and a task clock at event rate (the
tactile-punctuated scheduler of \S\ref{sec:punctuation}).

\subsubsection{Streaming and deployment}
The rolling cache commits in the frame-id causal order of training, with an attention
window aligned to the training window. Following observation-grounded streaming of the video
stream, we extend the same grounding to \emph{touch}: as real frames \emph{and}
real tactile readings arrive, they replace the corresponding predicted entries, so the
rolling context stays anchored to what the robot actually saw and felt on one timeline.
The observed tactile pathway of Eq.~(\ref{eq:local}) sits outside the denoising loop, so its
conditioning can be refreshed at sensor rate between chunk-level re-predictions. The first chunk has no committed history, so its first frame is
clamped to the observed initial frame, and the paired action is the zero delta of the
chunk-start anchor; alternatively, the first frame can be left free and denoised from noise like any other frame.

\subsubsection{Real-time inference}

Three properties of \mname{} enable its inherited rolling generation, caching mechanism, and asynchronous pipeline to operate in real time even at the $7$B-parameter scale:
(1)~The cascade imposes a one-way dependency within each action chunk: neither the video queries nor the tactile queries attend to the actions in the current chunk. Consequently, once the prediction phase has completed denoising the future video and tactile signals, their per-layer keys and values remain unchanged throughout the subsequent action loop. The asymmetric MoT architecture makes this strategy computationally effective: we cache these keys and values and \emph{only} re-run the lightweight action expert at each action denoising step.
(2)~The predicted video and tactile latents share one denoising-step schedule, advancing under the same scalar timestep at every step of the prediction loop, while the action uses an \emph{independent} schedule in its own loop. The action branch can therefore adopt a shorter denoising schedule without incurring the per-step computational cost of the video expert.
(3)~Across successive chunks, a rolling history cache reuses previously committed chunks, while the asynchronous pipeline predicts the next chunk as the robot executes the current one. As long as the computation for the next chunk fits within the execution window of the current chunk, the inference latency can be hidden behind robot execution.
As an optional deployment-time configuration, the model backbone can also be served at reduced precision. The linear layers in each expert operate in FP8, or NVFP4 on Blackwell-class hardware, while numerically sensitive operations (including the attention QKV projections and softmax, normalization layers, and output heads) remain in bf16. This strategy reduces the memory footprint and matrix-multiplication cost of the dominant video expert while having a negligible effect on the action output.
The additional computational cost introduced by the tactile modality is negligible. The predicted contact signal is generated in the same forward pass that already performs scene prediction, while the observed branch in Eq.~(\ref{eq:local}) consists only of a lightweight cross-attention module that is refreshed at sensor rate outside the denoising loop.

\subsubsection{Long-horizon execution via tactile punctuation}
\label{sec:punctuation}
\begin{figure}[H]
    \centering
    \includegraphics[width=\textwidth]{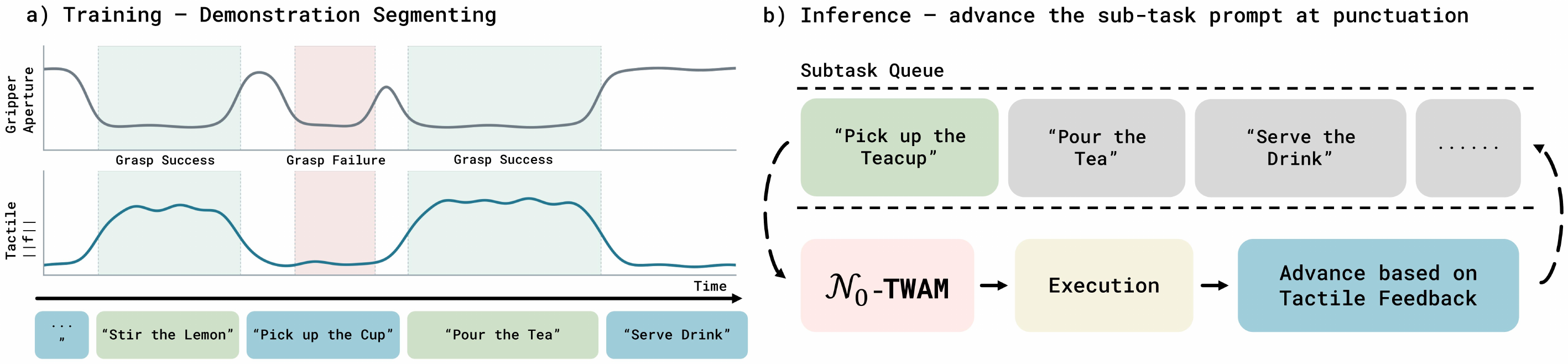}
    \caption{\textbf{Tactile-punctuated long-horizon execution.}
    \emph{(a)~Training:} a long-horizon demonstration, drawn as gripper aperture and
    tactile magnitude $\|f\|$ over time, is segmented at contact events into sub-task
    clips. The tactile trace also reports whether each grasp \emph{succeeds}: a gripper
    can close without loading any force, so a failed grasp shows an aperture dip but no
    tactile rise, and the one signal that delimits sub-tasks also confirms them. Each clip
    is labeled with a short instruction (``Pick up the Cup'', ``Pour the Tea'', \dots) and
    post-trained through the existing text pathway. \emph{(b)~Inference:} a lightweight
    scheduler holds the sub-task queue; \mname{} runs on the current sub-task prompt and
    executes it, and when tactile feedback signals the sub-task is complete the scheduler
    advances to the next prompt, looping through the queue.}
    \label{fig:punctuation}
\end{figure}

In long-horizon tasks, a single episode-level instruction does not tell the model \emph{which stage of the task} the robot is currently in. Mid-task observations are often stage-ambiguous (a table halfway through ``make a cup of tea''
looks much like the table before pouring), the attention window does not span the
episode, and a constant prompt carries no information about the current stage, so the
text pathway goes unused \emph{within} a task and the policy stalls, repeats a step, or
skips ahead. The missing structure already lives in the tactile stream: sub-tasks are
delimited by \emph{contact events} (a grasp begins at contact onset and ends at release;
a pour begins when the vessel is loaded and ends when it is set down), between which touch
is steady and at which it changes sharply. The tactile stream is naturally
\emph{punctuated}, and in \mname{} touch is already a stream the model observes
and predicts, so we let the same signal structure the task on both sides of training
(Fig.~\ref{fig:punctuation}).

\paragraph{Segmenting demonstrations.}
We segment long-horizon demonstrations using tactile ``punctuation,'' primarily detecting contact-onset and release events from changes in tactile signals,
with gripper aperture used as a fallback when necessary. The full demonstration is thereby divided into multiple sub-task clips.
This approach extends keyframe-based imitation learning~\cite{peract} to \emph{contact events}. Because a grasp produces a rise in tactile signals only after the gripper has actually made contact with an object, the same tactile reading can also distinguish a successful grasp from a gripper that closes on nothing (Fig.~\ref{fig:punctuation}a). Thus, a single signal is sufficient both to segment the demonstration and to verify task success. We use Gemini 3.5 Flash~\cite{gemini} to generate a short sub-task instruction for each clip, with human spot checks, and during post-training each clip is conditioned on its own corresponding sub-task instruction.

\paragraph{Advancing sub-tasks at inference.}
Touch is not only observed and predicted by the model, but also connected to the inference-time planner. At inference, a lightweight scheduler maintains a queue of sub-tasks and advances to the next prompt when the current sub-task completes (Fig.~\ref{fig:punctuation}b). Completion combines the model's two tactile roles: the predicted future tactile signal triggers the advance one step early, and the corresponding release or contact-onset event in the observed tactile stream then confirms it before the switch is committed. The sub-task sequence can be fixed in advance for repeatable tasks or generated by a vision--language planner in open-ended settings~\cite{saycan,codeaspolicies}; in either case, the tactile-event detector decides when to advance from the actual execution, and each advance re-seeds the streaming context with the next prompt, matching the per-clip conditioning used in training.

% ============================ 3. DATA ============================
\section{Data}
\label{sec:data}
\mname{} is trained on \emph{NeoData}~\cite{neodata}, and this section covers only what
the model consumes and how it is produced. We first summarize the slice of the dataset we
use (\S\ref{sec:data-stats}), then the offline pipeline that turns raw episodes into the
latent tokens the model operates on (\S\ref{sec:data-proc}).

\subsection{Dataset}
\label{sec:data-stats}
NeoData is a large multi-robot manipulation corpus of over $30{,}000$ hours, spanning
six embodiments and $450$ contact-rich tasks, in which a large fraction of the episodes
carry synchronized per-finger tactile alongside multi-view RGB. Tactile coverage at this
scale is what separates it from vision-only pre-training corpora, and is what lets
\mname{} learn to predict contact rather than only appearance. We refer to the NeoData
report~\cite{neodata} for the collection system, sensors, curation, and full statistics;
the real-robot post-training episodes use the InTac~S1 visuotactile sensors described in
\S\ref{sec:real_tasks}.

\subsection{Data processing}
\label{sec:data-proc}

\paragraph{Latent encoding.}
To train efficiently at the scale of $10^{9}$ frames, we encode both the camera views and
the tactile streams with a \emph{single} frozen causal video VAE~\cite{wan2024}, treating
each tactile stream as a small video so that touch shares the latent space of vision, and
we encode the language instruction with a frozen umT5 text encoder. All of these inputs are
converted to latents once during preparation and cached to disk, so the encoders never run
in the training forward pass.

\paragraph{Chunking.}
Each episode is segmented into fixed windows of $33$ latent frames, the unit over which the
chunked cascade of \S\ref{sec:arch} operates. A window spans $387$ raw frames at $30$\,fps,
subsampled to $129$ frames at $10$\,fps and compressed to $33$ latent frames by the VAE's
$4\times$ temporal downsampling. Consecutive windows advance by a stride of $24$ latent
frames, leaving a $\sim\!27\%$ overlap that keeps
segment junctions covered, and tail windows are right-aligned to the episode end. A
one-time validation pass records which windows are complete across all required modalities,
and only complete windows are used for training.

\paragraph{Unified action space.}
To train one policy across heterogeneous robots, every embodiment's action is
canonicalized into a single $20$-dimensional end-effector space, $10$ per arm. Every
embodiment is recorded with end-effector pose; the position and 6D-rotation targets are
formed as the chunk-anchored deltas of \S\ref{sec:arch} while the gripper is kept
absolute, and actions are normalized \emph{per robot} rather than globally, with quantile
clipping on the heavy-tailed gripper channel.

\paragraph{Contact-event stage segmentation.}
For the long-horizon pipeline of \S\ref{sec:punctuation} we also mark \emph{stage}
boundaries within a demonstration, with the \emph{tactile} stream as the primary signal:
most transitions in contact-rich tasks (first contact, loss of contact, slip, an insertion
seating) show up as a sharp change in the tactile reading. Where the tactile cue is weak or
ambiguous, we fall back to the gripper aperture, which reliably marks grasps and releases,
and a final human check corrects the remaining boundaries. This keeps the sub-task labeling
and inference-time scheduling of \S\ref{sec:punctuation} keyed on meaningful contact events.

% ===================== 5. EXPERIMENTS =====================
\section{Training and Experiments}
\label{sec:exp}
% =========================================================================
\subsection{Implementation Details}
\label{sec:impl}

\paragraph{Pre-training.}
We pre-train the world-action model on tens of thousands hours of manipulation, from the multi-robot corpus of \S\ref{sec:data}: real-robot
demonstrations across six embodiments (ARX5, UR, Flexiv, Franka, PiPER and hand-collected UMI) data. Each clip provides multi-view RGB pre-encoded with the Wan2.2 VAE
at $4\times$ temporal and $16\times$ spatial compression, a synchronized predicted tactile
stream, and the $20$-D bimanual delta end-effector action. The
training objective is the multi-modal flow-matching loss of Eq.~(\ref{eq:loss}): video,
touch, and action are trained with equal $1{:}1{:}1$ modality weights; every latent-frame
group is noised independently, with one noise level shared across a frame's spatial tokens
for video and tactile, and across the action steps aligned with that frame for action. At this
stage tactile enters \emph{only} as the predicted foresight target (\S\ref{sec:tactile}),
while the observed tactile pathway is disabled, so pre-training concentrates capacity on
predicting future vision and contact at scale. We follow a two-stage curriculum: the model
is first pre-trained on the real-robot data to convergence, then continued from that
checkpoint with UMI data mixed in (about $60\%$ of the batch). The video expert is a $\sim\!5$B video-diffusion
transformer of the Wan2.2-TI2V-5B family~\cite{wan2024}, warm-started from
LingBot-VA~\cite{lingbotva}, while the action and tactile experts are trained from scratch. We optimize with AdamW at a learning rate of $10^{-4}$ and weight decay $0.1$ for
$30{,}000$ steps, about $2.2$ epochs, at an effective batch of $512$. Training runs on
$128$ NVIDIA H800 GPUs under FSDP2 with bf16 mixed precision, and the optimizer updates
only the MoT backbone since the tokenizer and text encoder are frozen. We hold out $3\%$ of
each task's data as a validation split and score a random subset of it at each validation pass.

\paragraph{Tactile encoder (NeoForce).}
The observed tactile pathway (\S\ref{sec:tactile-local}) uses a ViT-B backbone initialized
from DINOv2~\cite{dinov2} and shared across the patchified visual and tactile streams, and each input is
a chunk of $4$ synchronized RGB frames and tactile force maps. We first pre-train NeoForce on
$20{,}000$ real-robot and Neo TacUMI demonstrations spanning $30$ tasks~\cite{neodata},
for $100{,}000$ steps on $8$ NVIDIA A100 GPUs at a learning rate of $2\times10^{-5}$.
\mname{} then loads that checkpoint and fine-tunes the encoder with the action loss during
post-training.

\paragraph{Post-training.}
For each downstream setting, we adapt the pre-trained \mname{} checkpoint to
task-specific demonstrations while preserving the multi-modal flow-matching cascade. Each
demonstration is segmented into synchronized windows of multi-view RGB, language, action,
and tactile, with RGB and language pre-encoded as Wan-VAE latents and text embeddings so
that optimization focuses on the transformer and tactile-conditioning modules. The action
is chunked at a horizon of $24$ steps, $12$ per latent frame. The
observed-tactile encoder differs by domain: on real robots it is the NeoForce force-space
encoder above, while in simulation, where the signal is not produced by a real sensor, it
is a lightweight encoder trained from scratch.

Observed tactile readings condition action prediction (\S\ref{sec:tactile}): the
current-frame tactile latent is embedded by the zero-initialized observed pathway and
cross-attended into the action expert under a chunk-causal mask, so these clean condition
tokens are visible to action queries, never enter the video stream, and are not themselves
a denoising target; the pretrained force-space head of \S\ref{sec:tactile} plugs into the
same port. In parallel, predicted tactile
latents remain a denoising target under the cascade, so the action expert draws on both
predicted future contact and observed current contact.

\paragraph{Condition dropout.}
During training we randomly drop the language and tactile conditions, each with
probability $0.1$, by two different mechanisms. Dropping language follows the standard
classifier-free recipe, replacing the text embedding with a learned null embedding.
Dropping tactile instead uses \emph{absence} rather than zeroing: on a dropped sample no
tactile tensor is passed at all, so the tactile tokens do not appear in the attention
sequence, the tactile modules receive no gradient, and that step trains the pure
video--action model. This
absence design has three consequences. First, it unifies deliberate dropout with genuinely tactile-less
data: episodes from embodiments without tactile sensors train through exactly the
same interface. Second, the model learns the marginal, vision-only policy alongside
the tactile-conditional one, so if a sensor fails or is absent at deployment the
policy degrades to its vision-only self instead of reacting to an out-of-distribution
zero input. Third, it exposes a classifier-free-guidance knob at inference: the gap
between the tactile-conditional and unconditional action predictions measures, and
can amplify, how much the policy listens to touch. In all reported evaluations we leave
this knob at $w{=}1$ (no tactile CFG amplification); the vision-only fallback it enables is
probed in \S\ref{sec:generalization}.

\paragraph{Optimization.}
Post-training data use the LeRobot dataset format, and the real-robot configuration uses
three RGB cameras and up to four tactile streams. The objective sums the video-latent, predicted-tactile, and
masked-action denoising losses at equal $1{:}1{:}1$ weights. We fine-tune all transformer
parameters, including the observed tactile encoder, with AdamW at a learning rate of
$10^{-4}$ and weight decay $0.1$, bf16 mixed precision, and gradient clipping at $2.0$;
task-specific runs use one sample per GPU, four-step gradient accumulation, and $1000$
optimizer steps unless noted.

% =========================================================================
\subsection{Evaluation Protocol}
\label{sec:real_tasks}
\begin{table}[t]
\centering
\caption{Success rate (\%) on the UniVTAC simulation benchmark (eight tasks).
Methods are grouped into vision-language-action (VLA) policies and world-action
models (WAM); the best per column is in \textbf{bold}.}
\label{tab:univtac}
\small
\renewcommand{\arraystretch}{1.25}
\begin{adjustbox}{width=\textwidth}
\begin{tabular}{lccccccccc}
\toprule
Method & \shortstack{Insert\\HDMI} & \shortstack{Insert\\Hole} & \shortstack{Insert\\Tube} & \shortstack{Grasp\\Classify} & \shortstack{Lift\\Can} & \shortstack{Lift\\Bottle} & \shortstack{Pull-out\\Key} & \shortstack{Put Bottle\\in Shelf} & Avg. \\
\midrule
\multicolumn{10}{l}{\emph{Vision-language-action (VLA)}} \\
\cmidrule(lr){1-10}
$\pi_{0.5}$~\cite{pi05} & 8 & 25 & 74 & 49 & 6 & \textbf{100} & 35 & 34 & 41.4 \\
StarVLA-$\alpha$~\cite{ye2026starvla} & 24 & 52 & 69 & 68 & 65 & 32 & 51 & \textbf{88} & 56.1 \\
InternVLA-A1~\cite{cai2026internvla} & 12 & 93 & 95 & 86 & 90 & 58 & 66 & 37 & 67.1 \\
Xiaomi-Robotics-0~\cite{cai2026xiaomirobotics} & \textbf{69} & 96 & \textbf{98} & 45 & 13 & 21 & \textbf{80} & 12 & 54.3 \\
\midrule
\multicolumn{10}{l}{\emph{World-action models (WAM)}} \\
\cmidrule(lr){1-10}
GigaWorld-Policy~\cite{gigaworld} & 0 & 12 & 9 & 20 & 0 & 38 & 32 & 21 & 16.5 \\
LingBot-VA~\cite{lingbotva} & 38 & 42 & 96 & 17 & 0 & 0 & 58 & 0 & 31.4 \\
FastWAM~\cite{fastwam} & 19 & 66 & \textbf{98} & 72 & 0 & 21 & 73 & 35 & 48.0 \\
\mname{}~(Ours) & 68 & \textbf{99} & \textbf{98} & \textbf{94} & \textbf{93} & 58 & 79 & 87 & \textbf{84.5} \\
\bottomrule
\end{tabular}
\end{adjustbox}
\end{table}

\paragraph{Benchmarks and metric.}
We evaluate on three suites: UniVTAC~\cite{chen2026univtac}, a public tactile-manipulation
simulation benchmark (eight tasks); NeoSim~\cite{neodata}, our in-house simulation suite
(twelve tasks, four single-arm and eight dual-arm); and a real-robot suite of eight
real-world contact-rich manipulation tasks, four on the dual-arm PiPER and four on the
single-arm Flexiv, six of which are drawn from NeoReal~\cite{neodata}. We report task success rate (\%) and its task average (a macro
average, weighting every task equally regardless of trial count); the best per column is
\textbf{bold}. Each simulation task is evaluated over $100$ trials and each real-robot
task over $20$ trials, every trial starting from a randomized initial configuration; the
reported success rate is the fraction of successful trials, with success given by the
benchmark-defined completion check in simulation and by a fixed per-task criterion judged
by the operator on the real robots.

\paragraph{Closed-loop action decoding.}
All methods are evaluated closed-loop under the asynchronous receding-horizon protocol of
\S\ref{sec:inference}. At each control cycle the policy consumes the current multi-view RGB
together with the rolling history cache and the current observed-tactile latent, runs the
cascade once to predict the future scene and contact, and denoises a single action
chunk---$16$ steps of the unified $20$-D bimanual delta end-effector action---with the slim
action expert on its own short denoising schedule. Chunks are executed asynchronously, the
robot running the current chunk while the next is predicted, and the observed-tactile
condition is refreshed at sensor rate between chunk-level re-predictions; a trial
terminates on task success or when it reaches a per-task step budget of $1.5\times$ the
length of that task's training demonstrations.

\paragraph{Baselines.}
We compare \mname{} against two kinds of baselines: world-action models and recent strong
vision-language-action (VLA) models. The world-action baselines are LingBot-VA, the unified video
world-action backbone our model builds on~\cite{lingbotva}; FastWAM, a
MoT world-action model~\cite{fastwam}; and GigaWorld-Policy, an
efficient action-centered world-action model~\cite{gigaworld}, all vision-only without
tactile. The VLA baselines are $\pi_{0.5}$~\cite{pi05},
StarVLA-$\alpha$~\cite{ye2026starvla}, InternVLA-A1~\cite{cai2026internvla}, and
Xiaomi-Robotics-0~\cite{cai2026xiaomirobotics}. Coverage varies by suite and each figure
and table lists the methods it includes: $\pi_{0.5}$, the strongest VLA, is run on all
three suites.

\begin{figure}[t]
    \centering
    \includegraphics[width=\textwidth]{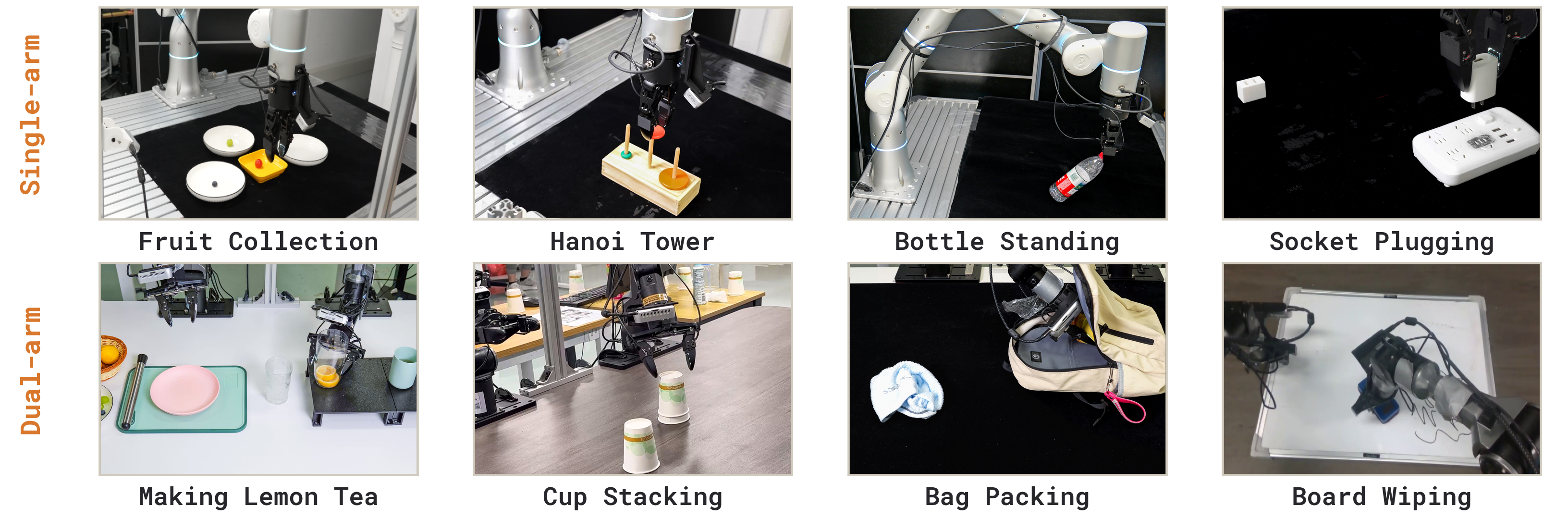}
    \caption{\textbf{The real-robot task suite.} Eight contact-rich tasks on two
    embodiments: four single-arm tasks on a Flexiv (top row) and four dual-arm tasks on a
    PiPER (bottom row). Six of the eight are drawn from NeoReal~\cite{neodata}.}
    \label{fig:real_tasks}
\end{figure}

\paragraph{Real-robot tasks and sensor.}
All real-robot experiments use the \textbf{InTac~S1} visuotactile sensor (Shanghai Xinzhi
Embodied Intelligence Technology Co., Ltd.\ / NeoteAI), mounted on two embodiments: a
dual-arm PiPER and a single-arm Flexiv. Our post-training suite has eight contact-rich
tasks (Fig.~\ref{fig:real_tasks}), each with tactile-critical decision points (secure grasp, incipient slip, first
contact, jamming, support transfer, final seating) that external cameras often cannot
resolve. The dual-arm PiPER runs Cup Stacking, Board Wiping,
Making Lemon Tea, and Bag Packing; the single-arm Flexiv runs
Fruit Collection, Bottle Standing, Socket Plugging, and
Hanoi Tower. Making Lemon Tea and Hanoi Tower are long-horizon tasks
(\S\ref{sec:punctuation}).

% =========================================================================
\subsection{Main Results}
\label{sec:main_results}

Following the evaluation protocol of \S\ref{sec:real_tasks} (benchmarks, metric, and
baselines), we report per-benchmark results below, with the best per column in
\textbf{bold}.

\paragraph{Simulation on UniVTAC.}
On the public UniVTAC tactile-manipulation benchmark (Table~\ref{tab:univtac})
we compare against vision-language-action (VLA) policies and world-action models (WAM).

\mname{} reaches the highest average success ($84.5\%$), about $17$ points above the
strongest baseline of any kind (InternVLA-A1, $67.1\%$). Notably, the vision-only
world-action baselines trail even the VLA policies here (LingBot-VA $31.4$, FastWAM
$48.0$, GigaWorld-Policy $16.5$): predicting the future scene is not enough for tasks
defined by contact. Making tactile native to the world-action model is what closes that
gap, lifting \mname{} past the VLA policies and the vision-only world-action baselines alike.

\paragraph{Simulation on NeoSim.}
Table~\ref{tab:neosim} reports per-task results on NeoSim, our in-house
simulation suite.

\begin{table}[H]
\centering
\caption{\textbf{Per-task success rate (\%) on the NeoSim simulation suite} (four
single-arm and eight dual-arm tasks). Methods are grouped into vision-language-action
(VLA) policies and world-action models (WAM); the best per row is in \textbf{bold}. FastWAM
and the ACT baseline were not run on NeoSim.}
\label{tab:neosim}
\small
\setlength{\tabcolsep}{4pt}
\renewcommand{\arraystretch}{1.15}
\begin{adjustbox}{width=\textwidth}
\begin{tabular}{l cccc ccc}
\toprule
 & \multicolumn{4}{c}{Vision-language-action (VLA)} & \multicolumn{3}{c}{World-action models (WAM)} \\
\cmidrule(lr){2-5}\cmidrule(lr){6-8}
Task & $\pi_{0.5}$ & StarVLA-$\alpha$ & InternVLA-A1 & Xiaomi-Robotics-0 & GigaWorld-Policy & LingBot-VA & \mname{} \\
\midrule
\multicolumn{8}{c}{\emph{Single-arm}} \\
\midrule
Insert USB              & 74 & 72 & 36 & 62 & 34 & 31 & \textbf{100} \\
Grasp Chip              & 86 & \textbf{93} & 12 & 49 & 71 & 77 & 92 \\
Unplug \& Plug Charger  & \textbf{23} & 0 & 0 & 5 & 3 & 0 & 0 \\
Pour Ball               & \textbf{92} & 32 & 47 & 52 & 0 & 37 & 63 \\
\midrule
\multicolumn{8}{c}{\emph{Dual-arm}} \\
\midrule
Bowl Unstack            & 3 & 17 & 0 & 0 & 0 & 18 & \textbf{72} \\
Cup Stack               & \textbf{22} & 0 & 0 & 0 & 12 & 20 & 12 \\
Insert Screw            & 26 & 8 & 0 & 0 & 0 & 18 & \textbf{29} \\
Place Gears             & \textbf{18} & 0 & 0 & 0 & 0 & 0 & 0 \\
Cup Handover            & 13 & \textbf{25} & 0 & 0 & 0 & 0 & 14 \\
Plate Stack             & 96 & 0 & 0 & 0 & 0 & 93 & \textbf{98} \\
Bowl Stack              & 93 & 0 & 8 & 42 & 0 & 67 & \textbf{97} \\
Cup Unstack             & 3 & 31 & 0 & \textbf{71} & 9 & 24 & 16 \\
\midrule
Average                 & 45.8 & 23.2 & 8.6 & 23.4 & 10.8 & 32.1 & \textbf{49.4} \\
\bottomrule
\end{tabular}
\end{adjustbox}
\end{table}

On NeoSim, \mname{} ($49.4$ average) is the best method overall, ahead of the
strongest baseline $\pi_{0.5}$ ($45.8$), and leads by a clear margin on the contact-heavy
stacking and insertion tasks (Insert USB, Bowl Unstack, Plate Stack, Bowl Stack). The tactile advantage is smaller here than on real
robots: the simulated tactile is not the sensor \mname{} was pretrained on, so its observed
pathway falls back to the lightweight sim encoder rather than the NeoForce force-space
representation (\S\ref{sec:tactile-local}). The real-robot results below are where
native touch matters most.

\begin{figure}[t]
    \centering
    \includegraphics[width=\textwidth]{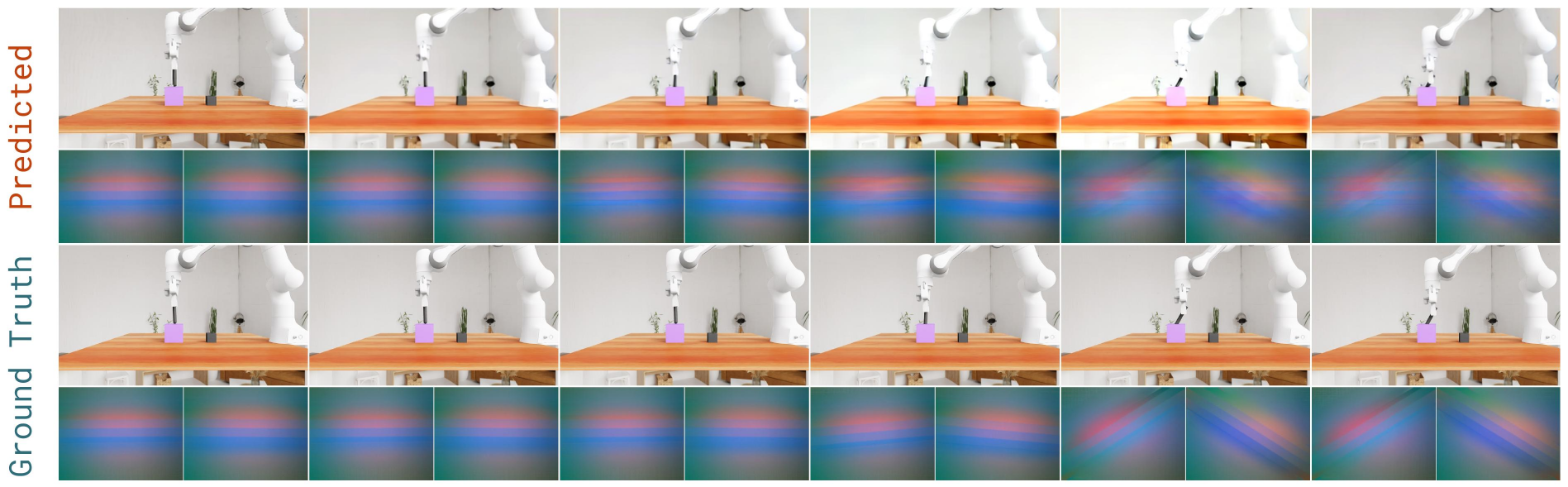}
    \caption{\textbf{Qualitative simulation rollout} (Insert Hole): the model's
    \emph{predicted} future (top) against the \emph{ground-truth} rollout (bottom). In each
    block the multi-view RGB scene sits above its tactile stream; the predicted scene and
    contact track the real execution across the insertion.}
    \label{fig:sim_demo}
\end{figure}

\begin{figure}[t]
    \centering
    \definecolor{nrPi}{HTML}{F8CE93}\definecolor{nrLg}{HTML}{DFA2D6}%
    \definecolor{nrFw}{HTML}{F4D9D2}\definecolor{nrOu}{HTML}{9CCBD7}%
    \definecolor{nrEdge}{HTML}{B9BFC4}\definecolor{nrInk}{HTML}{2F6D7A}%
    \begin{tikzpicture}[font=\footnotesize\rmfont]
      \foreach \c/\t/\x in {nrPi/{$\pi_{0.5}$}/0, nrLg/{LingBot-VA}/1.7,
                            nrFw/{FastWAM}/4.4, nrOu/{\mname{}}/6.7}{%
        \fill[\c,draw=nrEdge,line width=0.2pt] (\x,0) rectangle ++(0.26,0.26);
        \node[anchor=west,inner sep=2pt] at (\x+0.30,0.12) {\t};
      }
    \end{tikzpicture}\\[2pt]
    \begin{tikzpicture}[x=1cm,y=1cm,
        vlab/.style={font=\fontsize{5}{6}\selectfont\rmfont, above, inner sep=1pt},
        tlab/.style={font=\fontsize{6.5}{7.5}\selectfont\rmfont, text=black!78, below, inner sep=2.5pt, align=center}]
      \def\ph{3.2}\def\hbw{0.16}
      \foreach \gy in {20,40,60,80}{\draw[gray!16] (-0.7,\gy/80*\ph) -- (14.0,\gy/80*\ph);
        \node[font=\fontsize{5.5}{6.5}\selectfont\rmfont,gray!45!black,left,inner sep=2pt] at (-0.7,\gy/80*\ph) {\gy};}
      \node[font=\fontsize{5.5}{6.5}\selectfont\rmfont,gray!45!black,left,inner sep=2pt] at (-0.7,0) {0};
      \draw[gray!55] (-0.7,0) -- (14.0,0);
      \node[rotate=90,font=\fontsize{6.5}{7.5}\selectfont\rmfont,anchor=south] at (-1.18,\ph/2) {Success rate (\%)};
      \foreach \gx/\va/\vb/\vc/\vd/\lbl in {%
        0/50/40/20/60/{Fruit\\Collection}, 1.9/15/25/0/40/{Hanoi\\Tower},
        3.8/0/20/0/70/{Bottle\\Standing}, 5.7/60/10/20/70/{Socket\\Plugging},
        7.6/5/0/0/15/{Making\\Lemon Tea}, 9.5/50/30/25/45/{Cup\\Stacking},
        11.4/20/10/15/15/{Bag\\Packing}, 13.3/40/40/35/55/{Board\\Wiping}}{%
        \fill[nrPi,draw=nrEdge,line width=0.2pt] (\gx-0.54-\hbw,0) rectangle (\gx-0.54+\hbw,\va/80*\ph);
        \fill[nrLg,draw=nrEdge,line width=0.2pt] (\gx-0.18-\hbw,0) rectangle (\gx-0.18+\hbw,\vb/80*\ph);
        \fill[nrFw,draw=nrEdge,line width=0.2pt] (\gx+0.18-\hbw,0) rectangle (\gx+0.18+\hbw,\vc/80*\ph);
        \fill[nrOu,draw=nrEdge,line width=0.2pt] (\gx+0.54-\hbw,0) rectangle (\gx+0.54+\hbw,\vd/80*\ph);
        \node[vlab] at (\gx-0.54,\va/80*\ph) {\va};
        \node[vlab] at (\gx-0.18,\vb/80*\ph) {\vb};
        \node[vlab] at (\gx+0.18,\vc/80*\ph) {\vc};
        \node[vlab] at (\gx+0.54,\vd/80*\ph) {\vd};
        \node[tlab] at (\gx,-0.05) {\lbl};
      }
    \end{tikzpicture}
    \caption{\textbf{Per-task success rate (\%) on the real-robot suite}, eight
    contact-rich tasks, four methods each (bar value on top; \mname{} in blue). The
    methods are the vision-language-action policy $\pi_{0.5}$~\cite{pi05} and the
    world-action models LingBot-VA~\cite{lingbotva} and FastWAM~\cite{fastwam}. Macro
    average (equal weight per task): \mname{} $46.3$, $\pi_{0.5}$ $30.0$, LingBot-VA
    $21.9$, FastWAM $14.4$. Six of the eight tasks are drawn from the NeoReal
    benchmark~\cite{neodata}.}
    \label{fig:neoreal}
\end{figure}
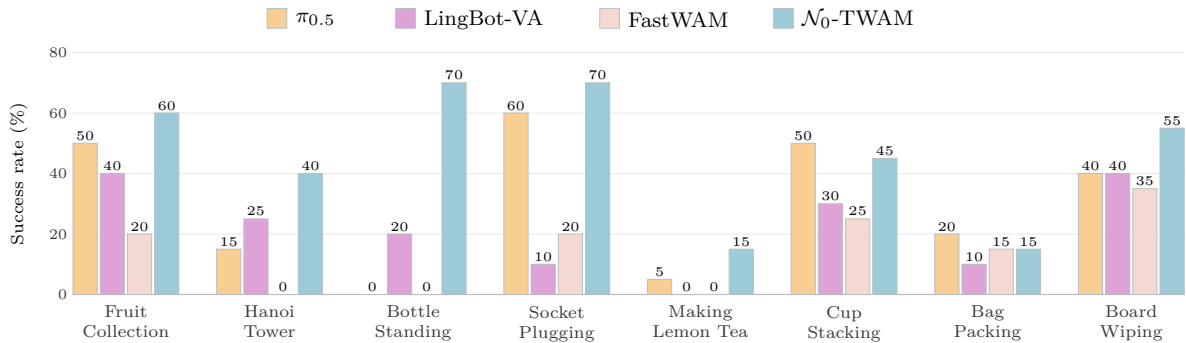

\paragraph{Real-world results.}
Figure~\ref{fig:neoreal} reports task-level success on our eight-task real-robot suite, six of them drawn from NeoReal. We compare against $\pi_{0.5}$, the strongest VLA policy
in our simulation experiments, and the world-action baselines LingBot-VA and FastWAM.

\paragraph{Overall ordering.}
Across the eight real-robot tasks, \mname{} is best overall, with a macro average of
$46.3$\% against $30.0$\% for $\pi_{0.5}$, $21.9$\% for LingBot-VA, and $14.4$\% for
FastWAM. $\pi_{0.5}$ is the strongest baseline, and LingBot-VA, which retains the full
video world-action prior, comes next, while the efficiency-oriented FastWAM trails both.
The margin is not uniform. It is largest exactly where success hinges on a hidden contact
state a camera cannot resolve: standing a bottle upright by observing its weight shift
(Bottle Standing, $70$ vs.\ $20$ for the best baseline), seating a plug (Socket Plugging,
$70$ vs.\ $60$), a secure fine grasp (Fruit Collection, $60$ vs.\ $50$), and wiping under
sustained contact load (Board Wiping, $55$ vs.\ $40$). On the two
tasks whose outcome is set by visually guided placement rather than contact, $\pi_{0.5}$
matches or edges \mname{} (Cup Stacking, $45$ vs.\ $50$; Bag Packing, $15$ vs.\ $20$),
consistent with the gain coming from touch rather than from task length alone.
Each real-robot number is a success rate over $20$ trials, so a per-task rate carries a
binomial standard error of up to $\sim\!11\%$; we therefore read individual per-task gaps
as indicative and rely on the consistent direction across tasks and the task-averaged margin.

\paragraph{Why \mname{} leads.}
\mname{} retains the strong video--action prior, but
adds touch to the control loop in two complementary forms. The predicted tactile expert
co-denoises future contact with future video, and the causal cascade lets the action
expert read this committed tactile future before producing the next action
(\S\ref{sec:tactile}). This anticipatory path helps the policy avoid actions whose
visual outcome appears plausible but whose predicted contact is inconsistent with a
stable grasp or insertion. In parallel, the observed pathway injects the current
force-space tactile representation directly into the action expert
(\S\ref{sec:tactile-local}), providing a short-latency correction signal for slip,
shear, asymmetric contact, and excessive pressure. The two paths therefore address
different failure modes: tactile foresight reduces errors before contact is committed,
whereas observed touch supports recovery after contact begins.

The implementation preserves the pretrained policy while adding these signals. Touch
uses private expert weights and exchanges information with vision and action only
through shared attention, preventing sparse, event-driven tactile statistics from
overwriting the visual representation. The staged recipe first learns predicted tactile
foresight at scale and activates the observed tactile reflex only during task adaptation,
preventing the reactive signal from becoming a shortcut that replaces predictive
contact modeling. The observed cross-attention output is zero-initialized, so post-training
starts from the pretrained behavior rather than an abruptly perturbed policy. Finally,
tactile-condition dropout trains the same action expert both with and without tactile
tokens, which limits over-reliance on touch during free-space motion and makes the
policy robust to temporarily missing contact observations. These design choices let
post-training add contact sensitivity without sacrificing the scene understanding
inherited from the base world-action model.

\paragraph{Task-level gains.}
The same mechanism explains where the advantage is largest. For Fruit Collection and
Bottle Standing, tactile feedback balances secure acquisition against object deformation,
slip, and a shifting center of mass. Board Wiping exposes sustained load and support
transfer against the surface. Socket Plugging and Hanoi Tower place the strongest demand
on both pathways: observed force direction supplies a correction signal after rim or peg
contact, while tactile foresight helps distinguish a recoverable alignment from an
impending jam or unstable seating. In the long-horizon tasks (Making Lemon Tea and Hanoi
Tower), these contact events also provide reliable boundaries for the tactile-punctuated
execution mechanism of \S\ref{sec:punctuation}, reducing the chance that one failed
transition corrupts the remaining action sequence. The two tasks where touch adds least,
Cup Stacking and Bag Packing, are also the two whose outcome is set by visually guided
placement, which is why the vision-strong $\pi_{0.5}$ stays on par or ahead there.

\paragraph{Baseline analysis.}
$\pi_{0.5}$ is the strongest baseline. As a large vision-language-action policy it brings
broad semantic priors and dependable visually guided placement, which keeps it close to
\mname{} on placement-dominated tasks and ahead on Cup Stacking and Bag Packing; but it
regresses actions directly from the current frame, with no model of the next moment and no
touch, so it can neither anticipate nor feel a contact event. Among the world-action
models, LingBot-VA ranks above FastWAM: it shares the strong visual foundation and retains
explicit future-video generation for reliable scene-level planning, whereas FastWAM's
efficiency-oriented inference collapses the video branch into a single-pass encoder rather
than preserving the iterative future rollout~\cite{fastwam}. None of the three baselines
has a tactile pathway, so visually similar contact states (``aligned but not inserted''
versus ``fully seated'') stay hard to tell apart. The ordering thus points to two
cumulative benefits: explicit iterative prediction separates LingBot-VA from FastWAM, and
adding both predicted and observed touch inside that prediction--action loop is what
carries \mname{} past every baseline, including the vision-strong $\pi_{0.5}$. The
component ablations in Figure~\ref{fig:ablation} test this attribution separately from the
aggregate baseline comparison.

% =========================================================================
\subsection{Generalization}
\label{sec:generalization}
We probe generalization on our real-robot suite along three axes, each isolated on the
task where it matters most (Table~\ref{tab:general}). \emph{Unseen object}: on Bottle
Standing and Bag Packing we keep the task and workspace fixed but replace the manipulated
objects with instances never seen in training, bottles of new shape, size, and material,
and new items to pack, so success requires transferring the contact skill to novel
geometry and physical properties rather than memorizing specific instances.
\emph{Unseen position}: on Cup Stacking we use the same cups but initialize them at
placements and offsets outside the training distribution, testing whether the stacking
behavior generalizes spatially rather than replaying a fixed trajectory.
\emph{Visual perturbation}: on Fruit Collection we leave the task and physics unchanged
but perturb the visual scene at test time (lighting and background changes),
stressing the visual pathway. Because tactile reports contact directly, \mname{} should
lean on touch and degrade more gracefully than vision-only world-action models as the
visual input shifts.

\begin{table}[H]
\centering
\caption{Generalization success rate (\%) on our real-robot suite; the best per column is
in \textbf{bold}.}
\label{tab:general}
\small
\begin{tabular}{lcccc}
\toprule
Method & Unseen Object & Unseen Position & Visual Perturbation & Avg. \\
\midrule
$\pi_{0.5}$~\cite{pi05}     & \textbf{80} & \textbf{45} & 25 & 50.0 \\
LingBot-VA~\cite{lingbotva} & 75 & 35 & 30 & 46.7 \\
\mname{}~(Ours)            & 65 & \textbf{45} & \textbf{45} & \textbf{51.7} \\
\bottomrule
\end{tabular}
\end{table}

The three axes separate cleanly. On \emph{unseen objects} $\pi_{0.5}$ leads ($80$): its
large vision--language pretraining carries broad object and semantic priors, so it adapts
to new instances best, whereas the world-action models lean on learned dynamics and
transfer less to novel appearance. On \emph{unseen positions} \mname{} matches $\pi_{0.5}$
(both $45$) and both clear LingBot-VA. The decisive axis is \emph{visual perturbation}: as
lighting and background shift, $\pi_{0.5}$ and LingBot-VA fall to $25$ and $30$, but
\mname{} holds at $45$, because tactile reports contact directly and gives the policy a
vision-independent signal to fall back on. \mname{} is therefore not the strongest on every
axis but the most \emph{robust}: it is highest on average ($51.7$) and degrades most
gracefully exactly where vision becomes unreliable, the regime touch is meant to cover.

% =========================================================================
\subsection{Ablation Study}
\label{sec:ablations}
We ablate the main design choices of \mname{}: the scale of the pre-training data, the
\emph{predicted} tactile foresight target, and the \emph{observed} tactile conditioning of the
action expert. The \emph{$20\%$ pre-training data} variant matches the full model in
architecture, task demonstrations, and post-training schedule, but warm-starts from a
checkpoint pre-trained on only a $20\%$ subset of the corpus rather than the full run,
isolating how much the downstream gains scale with pre-training data. The two tactile
variants remove one pathway each through a single switch, starting from the full model
that keeps both. The \emph{observed} pathway is a side branch that cross-attends the
current tactile into the action stream, so disabling it simply skips instantiating that
branch. The \emph{predicted} pathway is instead coupled into the backbone as both
self-attention segments and the
tactile expert's generation target, so rather than delete it we apply an
\emph{information ablation}: we zero the predicted tactile latent before it is noised,
leaving the sequence layout, attention mask, and diffusion loss token-for-token
identical to the full model and changing only the information the pathway carries. All
variants warm-start from the same checkpoint and share data, hyperparameters, and step
count. Each variant is evaluated on UniVTAC and NeoSim (the $20\%$ pre-training variant on
UniVTAC only); Figure~\ref{fig:ablation} reports the task-averaged success on each and their
mean, with the per-task breakdown in Appendix~\ref{app:ablation}. On UniVTAC, pre-training
scale is the largest single factor: warm-starting from a $20\%$ checkpoint lowers success
from $84.5$ to $65.4$. Both tactile pathways are load-bearing on top of that: removing the
predicted foresight target lowers UniVTAC success to $71.8$ and removing the observed
conditioning to $70.5$, and the NeoSim ablation shows the same ordering more sharply (from
$49.4$ down to $41.1$ and $29.6$), so touch contributes in both its predicted and its
observed form rather than through a single mechanism.

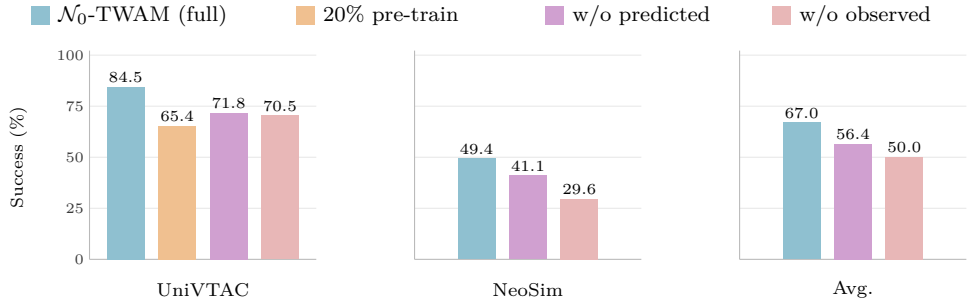
\begin{figure}[H]
\centering
\definecolor{abFull}{HTML}{8FC0D0}%
\definecolor{abPre}{HTML}{F0C090}%
\definecolor{abImg}{HTML}{D0A0D0}%
\definecolor{abSen}{HTML}{E8B7B7}%
% ---- shared legend (RobotoMono, well-spaced) ----
\begin{tikzpicture}[font=\footnotesize\rmfont]
  \foreach \c/\t/\x in {abFull/{\mname{} (full)}/0, abPre/{20\% pre-train}/3.5,
                        abImg/{w/o predicted}/6.8, abSen/{w/o observed}/9.8}{%
    \fill[\c] (\x,0) rectangle ++(0.24,0.24);
    \node[anchor=west, inner sep=2pt] at (\x+0.28,0.12) {\t};
  }
\end{tikzpicture}\\[4pt]
\begin{tikzpicture}[x=1cm,y=1cm, font=\scriptsize\rmfont]
  \def\pw{3.0}\def\ph{2.7}\def\bw{0.5}
  % panels: gridlines, axes, titles
  \foreach \ox/\title in {0/UniVTAC, 4.3/NeoSim, 8.6/{Avg.}}{%
    \foreach \gy in {25,50,75,100}{\draw[gray!18] (\ox,\gy/100*\ph) -- (\ox+\pw,\gy/100*\ph);}
    \draw[gray!55] (\ox,0) -- (\ox,\ph+0.06);
    \draw[gray!55] (\ox,0) -- (\ox+\pw,0);
    \node[font=\fontsize{7}{8}\selectfont\rmfont] at (\ox+\pw/2,-0.40) {\title};
  }
  % y-axis ticks + label (leftmost panel)
  \foreach \gy in {0,25,50,75,100}{\node[font=\fontsize{5.5}{6.5}\selectfont\rmfont,gray!45!black,left,inner sep=2pt] at (0,\gy/100*\ph) {\gy};}
  \node[rotate=90,font=\fontsize{6.5}{7.5}\selectfont\rmfont,anchor=south] at (-0.70,\ph/2) {Success (\%)};
  % bars packed left-to-right per panel (no empty slot): offset / slot-x / value / color
  \foreach \ox/\sx/\val/\col in {%
    0/0.48/84.5/abFull, 0/1.16/65.4/abPre, 0/1.84/71.8/abImg, 0/2.52/70.5/abSen,
    4.3/0.82/49.4/abFull, 4.3/1.50/41.1/abImg, 4.3/2.18/29.6/abSen,
    8.6/0.82/67.0/abFull, 8.6/1.50/56.4/abImg, 8.6/2.18/50.0/abSen}{%
      \fill[\col] (\ox+\sx-\bw/2,0) rectangle (\ox+\sx+\bw/2,\val/100*\ph);
      \node[font=\fontsize{5.8}{7}\selectfont\rmfont,above,inner sep=1.5pt] at (\ox+\sx,\val/100*\ph) {\val};
  }
\end{tikzpicture}
\caption{\textbf{Ablation of \mname{}} on UniVTAC (eight tasks), NeoSim (twelve tasks), and
their mean. Removing either tactile pathway lowers success, most sharply on NeoSim; the
$20\%$ pre-training variant (UniVTAC only) shows pre-training scale is the largest single
factor. Per-task breakdowns are in Appendix~\ref{app:ablation}.}
\label{fig:ablation}
\end{figure}

% =========================================================================
\subsection{Analysis}
\label{sec:analysis}

\paragraph{Delta vs.\ absolute end-effector actions.}
We default to the chunk-anchored \emph{delta} end-effector parameterization of
\S\ref{sec:arch}, which is translation-invariant and keeps the targets small and centered,
so it transfers cleanly across embodiments and workspace placements. It is not always the
best choice at deployment. On alignment-sensitive tasks the absolute end-effector pose can
outperform the delta target (Table~\ref{tab:delta_abs}): a delta is measured from the moving
chunk-start anchor, so small per-chunk errors accumulate and the target is never tied to a
fixed location, whereas an absolute pose grounds each target directly in the world frame.
The two parameterizations therefore trade generalization against precision: delta for
translation-invariant transfer, absolute for reaching a specific pose. We keep delta for the
unified multi-embodiment policy and adopt absolute on the task-specific settings where
sub-centimeter placement dominates success.

\begin{table}[h]
\centering
\caption{Delta vs.\ absolute end-effector parameterization on four alignment-sensitive
tasks (success rate \%). The absolute pose, which we adopt for these tasks, avoids the
drift the delta target accumulates.}
\label{tab:delta_abs}
\small
\begin{tabular}{lccccc}
\toprule
Parameterization & Lift Can & Lift Bottle & Put Bottle in Shelf & Grasp Chip & Avg. \\
\midrule
Delta      & $24$ & $52$ & $86$ & $38$ & $50.0$ \\
Absolute   & $\mathbf{93}$ & $\mathbf{58}$ & $\mathbf{87}$ & $\mathbf{92}$ & $\mathbf{82.5}$ \\
\bottomrule
\end{tabular}
\end{table}

\begin{figure}[t]
    \centering
    \definecolor{tlabink}{HTML}{2F6D7A}%
    \setlength{\tabcolsep}{2pt}%
    \newcommand*\tlab[2][1.5ex]{\raisebox{#1}{\smash{\rotatebox[origin=c]{90}{\color{tlabink}\fontfamily{RobotoMono-TLF}\selectfont\fontsize{6}{6.5}\selectfont #2}}}}%
    \begin{tabular}{@{}c@{\hspace{5pt}}c@{}}
        \tlab[3.3ex]{\shortstack{RGB\\scene}} &
        \includegraphics[width=0.95\textwidth]{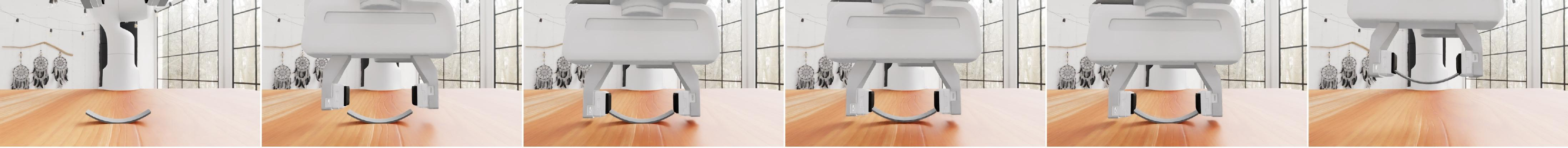} \\[3pt]
        \tlab[2.3ex]{Clean} &
        \includegraphics[width=0.95\textwidth]{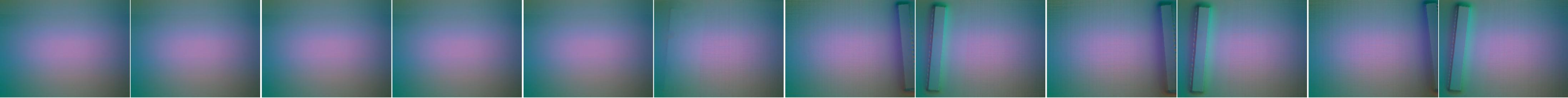} \\[3pt]
        \tlab[2.3ex]{\shortstack{Gel-\\rendered}} &
        \includegraphics[width=0.95\textwidth]{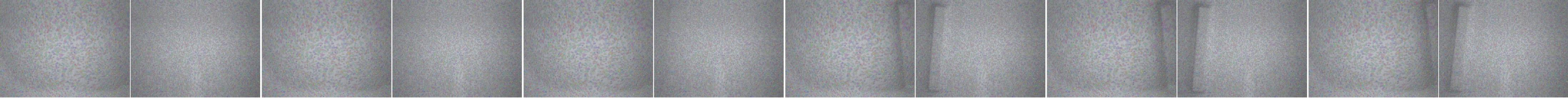} \\
    \end{tabular}
    \caption{\textbf{Two renderings of the same tactile stream.} One observed rollout: the
    top row is the RGB scene, shared by both renderings; below it the same contact appears
    as the simulator's smooth \emph{clean} field and as our \emph{gel-rendered}
    rendering, a silvery, speckled gel with grain noise and depth shading that matches the
    look of a real vision-based sensor. Only the tactile rendering differs.}
    \label{fig:tactile_realism}
\end{figure}

\paragraph{Realism of the simulated tactile image.}
Both tactile images in this study come from the \emph{same} contact simulation and differ
only in how the deformed gel is imaged. The UniVTAC numbers above use the simulator's
\emph{clean} output, a calibration-based GelSight render (Taxim~\cite{taxim}) that
maps the simulated contact-height field to a smooth, deterministic RGB image through a
per-pixel illumination model, so contact appears as a clean color-and-intensity blob. A
real vision-based sensor looks nothing like this; it returns a silvery, speckled, softly
shaded gel image with sensor noise (Fig.~\ref{fig:tactile_realism}). We therefore render a
more realistic \emph{gel-rendered} image on top of the same signal, without re-simulating:
we rebuild a bright silver gel base from the optical image's luminance, overlay a dense,
\emph{markerless} field of fine colored speckles, and \emph{advect} that field with the
contact, tangentially by the per-vertex surface displacement (the same field that drives
the marker motion) and normally by the depth-indentation gradient, which pushes the
speckles outward as the gel bulges around the indenter; a faint depth shade and elastomer
grain complete the look. Contact then reads out of a \emph{deforming textured surface}
rather than a color code. Trained and evaluated on this rendering, \mname{} stays close to the
clean field rather than dropping sharply: $82.4\%$ against $84.5\%$ on the eight UniVTAC tasks,
and $58.5\%$ against $63.8\%$ (single-arm) and $39.3\%$ against $42.3\%$ (dual-arm) on NeoSim
(Table~\ref{tab:realism_avg}; the per-task breakdown is in Appendix~\ref{app:realism}). The
realistic appearance therefore costs little accuracy in-domain, even though it replaces the
clean synthetic blob with sensor noise, grain, and a deforming textured surface. Because it
shares the visual domain of a real vision-based sensor, the gel-rendered image is also a tactile
stream a pretrained force encoder can read, which we examine below.

\begin{figure}[t]
    \centering
    \includegraphics[width=\textwidth]{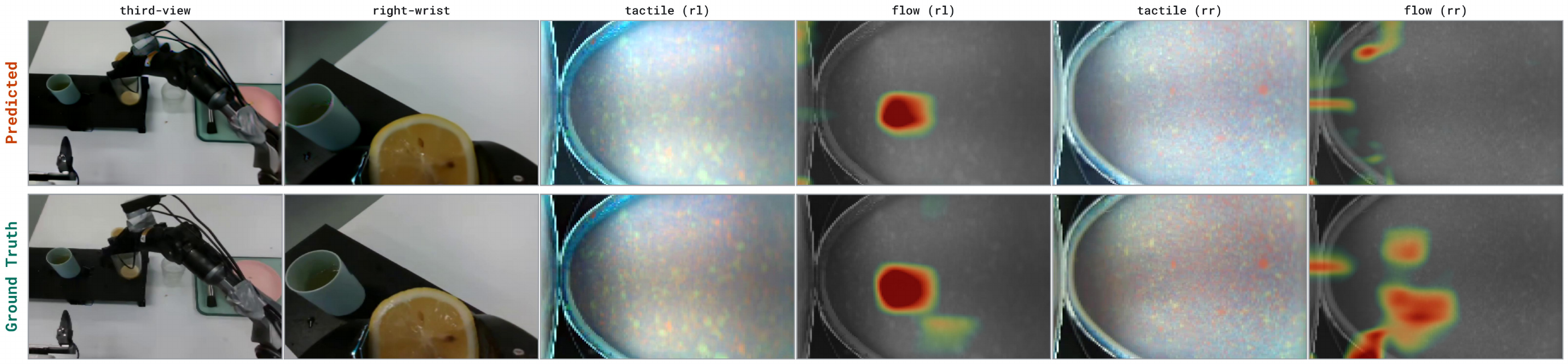}
    \caption{\textbf{Predicting future touch} (lemon-tea grasp). Top row: the model's
    \emph{predicted} future; bottom row: the \emph{ground truth}. From left to right, the
    third-person and right-wrist RGB views, then the tactile image and the contact-flow
    map for each of the two gripper fingers (\emph{rl}, \emph{rr}). The predicted tactile
    and flow match the observed sensor, so foresight extends to contact, not only to the
    visual scene.}
    \label{fig:foresight}
\end{figure}

\begin{table}[H]
\centering
\caption{Realism of the simulated tactile, and the pretrained NeoForce force encoder on it.
Average success (\%) for the simulator's clean field, our gel-rendered image, and the
gel-rendered image read through NeoForce, on UniVTAC (eight tasks) and NeoSim (four single-arm
and eight dual-arm tasks). Per-task numbers are in Appendix~\ref{app:realism}.}
\label{tab:realism_avg}
\begin{tabular}{lccc}
\toprule
Tactile & UniVTAC & NeoSim (single) & NeoSim (dual) \\
\midrule
Clean (simulator)        & $84.5$ & $63.8$ & $42.3$ \\
Gel-rendered             & $82.4$ & $58.5$ & $39.3$ \\
Gel-rendered + NeoForce  & $\mathbf{88.1}$ & $\mathbf{64.8}$ & / \\
\bottomrule
\end{tabular}
\end{table}

\paragraph{Pretrained force encoder on the gel-rendered tactile.}
NeoForce is pretrained on real vision-based tactile images, the gel-particle appearance our
gel-rendered render imitates. Because the two share this domain, reading the gel-rendered tactile
through NeoForce, instead of the observed-tactile encoder trained from scratch, raises success
from $82.4\%$ to $88.1\%$ on the eight UniVTAC tasks and from $58.5\%$ to $64.8\%$ on the four
single-arm NeoSim tasks (Table~\ref{tab:realism_avg}; per-task numbers in
Appendix~\ref{app:realism}). The gains concentrate on tasks whose success turns on a
weight-bearing or shear cue that a force representation reads well: Put Bottle in Shelf
($82\!\to\!96$), Insert HDMI ($63\!\to\!75$), and Pull-out Key ($74\!\to\!83$) on UniVTAC, and
Grasp Chip ($82\!\to\!92$) and Pour Ball ($55\!\to\!68$) on NeoSim, while the sharp-contact
insertion tasks are already near ceiling and change little. We still report the clean field in the
main results, where every method uses the same simulated tactile; the gel-rendered\,+\,NeoForce
number is a sim-to-real signal, showing that once the simulated touch looks real, a force encoder
pretrained on real touch transfers into it.

\paragraph{Tactile foresight.}
A direct test of whether the model truly \emph{predicts} touch is to compare its
predicted future tactile against the ground truth (Fig.~\ref{fig:foresight}). On the
lemon-tea grasp, the predicted tactile images and their contact-flow maps track the
ground-truth sensor closely for both fingers: the model anticipates where and how contact
forms, not only the future scene. This is the qualitative signal we expect if predicting
\emph{touch}, rather than appearance alone, is what drives the downstream gains.

% ============================ 2. RELATED WORK ============================
\section{Related Work}
\label{sec:related}

\paragraph{Vision-language-action models.}
Vision-language-action (VLA) models cast manipulation as conditional action generation
from observations and language. RT-style and open generalist
policies~\cite{rt1,rt2,octo,openvla,pi0,pi05} show that large-scale vision-language
pretraining transfers semantic knowledge into robot control. Recent systems extend this
recipe through real-time asynchronous execution and much larger cross-embodiment
corpora~\cite{cai2026xiaomirobotics,xiaomi2026robotics1,wu2026lingbotvla2}, while embodied
backbones and unified policies add action-centric reasoning and visual or latent
foresight~\cite{tencent2026hyembodied05,wang2026hyembodiedvlm,cai2026internvla,internvla_a15}.
Yet these foundation-scale systems still center on vision, language, and proprioception.
They are typically measured on unified sim-and-real benchmarks such as VLABench~\cite{vlabench}, LIBERO-Plus~\cite{liberoplus}, and RoboDojo~\cite{chen2026robodojo} that span generalization, memory, precision, and
long-horizon skills, but not the contact-rich, tactile-dependent tasks \mname{} targets.
In contact-rich manipulation, the deciding variable may instead be fingertip pressure,
incipient slip, or imminent collision; even predictive variants model visual or latent
futures rather than an explicit tactile trajectory consumed by the action expert.
\mname{} targets this gap.

\paragraph{Tactile perception for manipulation.}
Tactile perception supplies the contact signals vision misses. Tactile-aware VLA
systems~\cite{zhang2026touchguide,tactilevla,vtla,tla,forcevla} align tactile observations
with language and action~\cite{tactilevla,tla}, inject force feedback into the action
expert~\cite{forcevla}, specialize for insertion and other contact-heavy tasks~\cite{vtla},
or steer a pretrained visuomotor policy at inference with a touch-derived feasibility
signal~\cite{zhang2026touchguide}. Force-aware policies fuse RGB with high-rate
force/torque streams, learn task-dependent compliance, or add torque adapters to a VLA
decoder~\cite{foar,manipforce,momaforce,acp,tavla}, and visuo-tactile policies show that
touch improves fine-grained manipulation and stays robust when vision is ambiguous or
occluded~\cite{threedvitac,vital,vlatouch,omnivtla,threedtacdex,seeingtouch}. In this direct-policy
family touch is largely \emph{reactive}~\cite{trex}: it enters as an immediate observation for the next
action rather than a contact trajectory the model predicts before acting. A more recent
line makes touch \emph{predictive}: DreamTacVLA fine-tunes a tactile world model so actions
condition on both real and predicted tactile consequences~\cite{dreamtacvla}, predictive
visuo-tactile policies autoregress future contact or generate force-domain action
trajectories~\cite{vitacformer,tacdiffusion}, and visuo-tactile world models and tactile
action models study contact-aware prediction for planning and closed-loop
control~\cite{higuera2026vtwm,omnivta,tacforesight,contactworld,vtam,dreamtac,tactilewam,tian2026vtwam}.
Against this prior art the question \mname{} addresses is not whether predicted touch helps,
but how to keep it in the action loop at scale while preserving a pretrained visual world
model.

\paragraph{World models for robotics.}
World models learn a predictive model of the environment and roll out future states
conditioned on text~\cite{wan2024}, driving commands~\cite{gaia1}, or robot
actions~\cite{ivideogpt}; trained on large video corpora, world foundation
models~\cite{cosmos} generate high-fidelity future observations, recover
action-controllable dynamics from unlabelled video~\cite{genie}, or predict in a latent
space rather than pixels~\cite{vjepa2}. Rather than map the current observation directly to
an action, a complementary line first \emph{predicts} a future and then derives the action
from it, from model-based agents that optimize a policy against predicted
rollouts~\cite{dreamer,dreamerv2,planet,worldmodels,diamond} to video-first policies that
plan through predicted visual futures~\cite{unipi,vlp,avdc,dreamitate}. World-action models
(WAMs) couple the two stages in a single
model~\cite{gr1,pad,uva,uwm,worldvla,lingbotva,dreamzero}, sharing or repurposing visual
latent spaces for action~\cite{gr2,motus,repwam,cosmospolicy} under various temporal
abstractions~\cite{wallwm,cronusvla,beingh07}. To give each modality its own capacity while
keeping a shared reasoning path, MoT~\cite{mot} decouples
per-modality feed-forward, attention, and normalization but preserves a shared global
attention~\cite{lingbotvideo}; LingBot-VA, which \mname{} builds on, applies an MoT WAM over
vision and action tokens~\cite{lingbotva}, and causal video training improves long-horizon
rollouts~\cite{selfforcing,nextforcing}. Recent WAMs differ chiefly in how they pay for
future prediction at test time: FastWAM collapses the video branch into a single-pass
encoder~\cite{fastwam}, Metis lets the action expert bypass future video
generation~\cite{metis}, and Efficient-WAM prunes and shortens the video
rollout~\cite{efficientwam}, with related distillation and short-cutting
elsewhere~\cite{svam,flashwam,fastdvla}; across this line efficiency is usually bought by
weakening or bypassing the predicted future, and the predicted modality is pixels. \mname{}
occupies the tactile version of this space: it orders the video, tactile, and action
experts into a cascade so predicted touch is an intermediate prediction the action expert
consumes rather than a side signal or attention bias~\cite{tactilewam,tian2026vtwam}, and it
allocates width asymmetrically to keep full pretrained capacity in the visual expert while
training slimmer tactile and action experts. The same asymmetry keeps streaming inference
cheap without pruning the predicted future away.

% ============================ 6. CONCLUSION ============================
\section{Conclusion}
\label{sec:conclusion}

We presented \mname{}, a tactile-native world-model policy that brings touch into the
predicted future of a video world-action model. By organizing a pretrained
video-diffusion backbone into three per-modality experts tied by a single shared
self-attention, and ordering them with a frame-id causal cascade, \mname{} predicts
the near-future scene and the near-future contact and then reads action out of both.
Tactile is given a dual role (a denoised foresight target and an observed
reading), while an asymmetric design keeps the model at 7.2B and
streaming inference cheap, since each action-denoising step re-runs only the lightweight action expert. Crucially, touch is not an add-on evaluated on a handful of tasks: \mname{} is
pre-trained at scale, on a large self-collected real-robot
data with synchronized tactile, so that a shared tactile representation is learned
jointly with vision rather than fitted per task. Across contact-rich benchmarks it is the
strongest method overall, in simulation ($84.5$ on UniVTAC, $49.4$ on NeoSim) and on real
robots ($46.3\%$ average, ahead of every vision-language-action and vision-only world-action
baseline), with ablations confirming that both the predicted and the observed tactile pathways
contribute.

\paragraph{Future work.}
We see three main directions. First, \emph{faster inference}: streaming decoding of the
one-directional cascade can be accelerated further, so the policy runs at higher control
rates. Second, \emph{a longer predicted horizon}: extending the vision and tactile
prediction window would let the model anticipate contact events further ahead. Third,
\emph{broader sensor coverage}: training on more types and varieties of tactile sensors
would widen the range of embodiments the learned tactile representation transfers to.

\section*{Contributors}

\textbf{Pre-Training.} Li Kang, Xiufeng Song, Yanjun Li, Rui Li, Shunlin Lu, Yiran Qin.\\
\textbf{Post-Training (Simulation).} Yifan Wang, Li Kang, Zipei Ma, Bruno N.Y. Chen, Heng Zhou.\\
\textbf{Post-Training (Real Robot).} Li Kang, Yanjun Li, Zipei Ma, Shengqi Xu, Boyu Mi, Bruno N.Y.\ Chen, Heng Zhou, Ren Jia, Silong Dai, Jiongwei Lu.\\
\textbf{Data Processing.} Li Kang, Boyu Mi, Rui Li, Xiufeng Song, Yifan Wang, Yutao Fan, Longjie Su, Zhemeng Zhang, Xin Wang, Tianyu Yang, Wenjie Zhou.\\
\textbf{Academic Supervision.} Ziyi Ye, Guoxiang Dong, Xiaosong Jia, Wenming Chen. \\
\textbf{Project Lead.}  Yiran Qin, Shunlin Lu, Shihao Zhao, Daoguo Dong, Zuxuan Wu.

% ===========================================================================
\bibliography{main}

% ===========================================================================
\appendix
% ============================ APPENDIX ============================
\section{Per-task ablation results}
\label{app:ablation}
Table~\ref{tab:ablation_full} gives the per-task UniVTAC success rates behind the averages
in Figure~\ref{fig:ablation}, for the full model and the three ablations, and
Table~\ref{tab:ablation_neosim} gives the twelve-task NeoSim ablation (four single-arm and
eight dual-arm).

\begin{table}[H]
\centering
\caption{Per-task UniVTAC ablation of \mname{} (success rate \%) over the eight tasks.}
\label{tab:ablation_full}
\small
\begin{tabular}{lccccccccc}
\toprule
Variant & \shortstack{Insert\\HDMI} & \shortstack{Insert\\Hole} & \shortstack{Insert\\Tube} & \shortstack{Grasp\\Classify} & \shortstack{Lift\\Can} & \shortstack{Lift\\Bottle} & \shortstack{Pull-out\\Key} & \shortstack{Put Bottle\\in Shelf} & Avg. \\
\midrule
\mname{}~(full)        & 68 & 99 & 98 & 94 & 93 & 58 & 79 & 87 & 84.5 \\
\quad $20\%$ pre-train  & 52 & 78 & 96 & 95 & 80 & 51 & 25 & 46 & 65.4 \\
\quad w/o predicted      & 73 & 97 & 96 & 93 & 91 & 16 & 86 & 22 & 71.8 \\
\quad w/o observed        & 68 & 93 & 99 & 58 & 68 &  0 & 86 & 92 & 70.5 \\
\bottomrule
\end{tabular}
\end{table}

\begin{table}[H]
\centering
\caption{Per-task NeoSim ablation of \mname{} (success rate \%) over the twelve tasks
(four single-arm and eight dual-arm). The $20\%$ pre-training variant was run on UniVTAC only.}
\label{tab:ablation_neosim}
\scriptsize
\setlength{\tabcolsep}{3pt}
\begin{tabular}{l cccc cccccccc c}
\toprule
 & \multicolumn{4}{c}{Single-arm} & \multicolumn{8}{c}{Dual-arm} & \\
\cmidrule(lr){2-5}\cmidrule(lr){6-13}
Variant & \shortstack{Insert\\USB} & \shortstack{Grasp\\Chip} & \shortstack{Unplug \&\\Plug Charger} & \shortstack{Pour\\Ball} & \shortstack{Bowl\\Unstack} & \shortstack{Cup\\Stack} & \shortstack{Insert\\Screw} & \shortstack{Place\\Gears} & \shortstack{Cup\\Handover} & \shortstack{Plate\\Stack} & \shortstack{Bowl\\Stack} & \shortstack{Cup\\Unstack} & Avg. \\
\midrule
\mname{}~(full)   & 100 & 92 & 0 & 63 & 72 & 12 & 29 & 0 & 14 & 98 & 97 & 16 & 49.4 \\
\quad w/o predicted & 82 & 61 & 0 & 38 & 18 & 22 &  6 & 0 & 65 & 96 & 97 &  8 & 41.1 \\
\quad w/o observed   & 88 & 24 & 0 & 17 & 32 & 38 & 10 & 0 & 30 & 15 & 96 &  5 & 29.6 \\
\bottomrule
\end{tabular}
\end{table}

\section{Per-task tactile-realism results}
\label{app:realism}
Tables~\ref{tab:realism_full} and~\ref{tab:realism_neosim_full} give the per-task success
behind Table~\ref{tab:realism_avg}: \mname{} on the simulator's clean field, on our
gel-rendered image, and on the gel-rendered image read through the pretrained NeoForce force
encoder. The clean and gel-rendered results are close on most tasks, so the realistic
appearance costs little accuracy in-domain. Reading the gel-rendered image through NeoForce
improves it further, most on the tasks with a clear weight-bearing or shear cue
(Put Bottle in Shelf $82\!\to\!96$, Insert HDMI $63\!\to\!75$, Pull-out Key $74\!\to\!83$ on
UniVTAC; Grasp Chip $82\!\to\!92$, Pour Ball $55\!\to\!68$ on NeoSim).

\begin{table}[H]
\centering
\caption{Per-task success rate (\%) on the eight UniVTAC tasks for the clean field, the
gel-rendered image, and the gel-rendered image read through NeoForce.}
\label{tab:realism_full}
\small
\begin{tabular}{lccccccccc}
\toprule
Tactile & \shortstack{Insert\\HDMI} & \shortstack{Insert\\Hole} & \shortstack{Insert\\Tube} & \shortstack{Grasp\\Classify} & \shortstack{Lift\\Can} & \shortstack{Lift\\Bottle} & \shortstack{Pull-out\\Key} & \shortstack{Put Bottle\\in Shelf} & Avg. \\
\midrule
Clean                   & 68 & 99 & 98 & 94 & 93 & 58 & 79 & 87 & 84.5 \\
Gel-rendered            & 63 & 98 & 99 & 97 & 94 & 52 & 74 & 82 & 82.4 \\
Gel-rendered + NeoForce & 75 & 98 & 100 & 96 & 99 & 58 & 83 & 96 & 88.1 \\
\bottomrule
\end{tabular}
\end{table}

\begin{table}[H]
\centering
\caption{Per-task success rate (\%) on the twelve NeoSim tasks (four single-arm and
eight dual-arm) for the clean field, the gel-rendered image, and the gel-rendered image read
through NeoForce.}
\label{tab:realism_neosim_full}
\scriptsize
\setlength{\tabcolsep}{3pt}
\begin{tabular}{l cccc cccccccc c}
\toprule
 & \multicolumn{4}{c}{Single-arm} & \multicolumn{8}{c}{Dual-arm} & \\
\cmidrule(lr){2-5}\cmidrule(lr){6-13}
Tactile & \shortstack{Insert\\USB} & \shortstack{Grasp\\Chip} & \shortstack{Unplug \&\\Plug Charger} & \shortstack{Pour\\Ball} & \shortstack{Bowl\\Unstack} & \shortstack{Cup\\Stack} & \shortstack{Insert\\Screw} & \shortstack{Place\\Gears} & \shortstack{Cup\\Handover} & \shortstack{Plate\\Stack} & \shortstack{Bowl\\Stack} & \shortstack{Cup\\Unstack} & Avg. \\
\midrule
Clean                   & 100 & 92 & 0 & 63 & 72 & 12 & 29 & 0 & 14 & 98 & 97 & 16 & 49.4 \\
Gel-rendered            & 97 & 82 & 0 & 55 & 74 & 20 & 17 & 0 & 12 & 96 & 83 & 12 & 45.7 \\
Gel-rendered + NeoForce & 99 & 92 & 0 & 68 & / & / & / & / & / & / & / & / & / \\
\bottomrule
\end{tabular}
\end{table}

\section{Per-modality training dynamics}
\label{app:loss}
\begin{figure}[H]
    \centering
    \includegraphics[width=\textwidth]{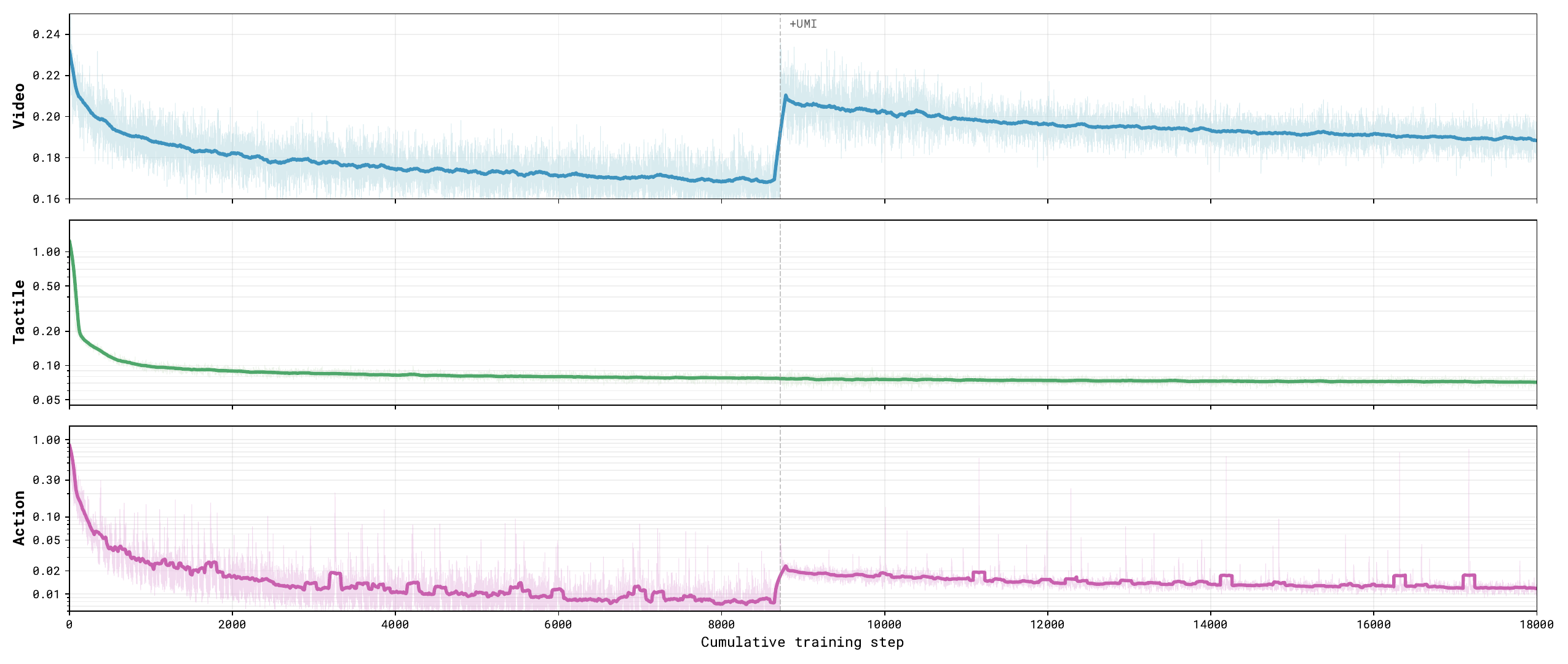}
    \caption{\textbf{Per-modality pre-training loss} (first 18k steps).
    Video, action, and tactile flow-matching losses of the MoT backbone;
    the \texttt{+UMI} marker is where the hand-collected UMI data is mixed in. The video loss
    rises and converges higher afterward, while the tactile loss is essentially unchanged.}
    \label{fig:mot_loss}
\end{figure}

We consider two pre-training data recipes: one that trains on real-robot teleoperation data
only, and one that additionally mixes in the hand-collected UMI data and then continues on
the pure real-robot teleoperation data. Figure~\ref{fig:mot_loss} shows the latter over the
first 18k steps (the full run is $30{,}000$ steps). Because the MoT keeps
a separate expert per modality, we can log the flow-matching loss separately for the video,
action, and tactile streams. Before the \texttt{+UMI} marker all three fall and stabilize.
When the UMI data is mixed in (about $60\%$ of the batch) the modalities respond
differently: the \emph{video} loss
rises at the transition and settles at a somewhat higher value than before, since the UMI
footage is a new visual distribution the video expert must now also fit; the \emph{tactile}
loss barely moves, because the tactile signal is similar across the two data sources; and
the \emph{action} loss shows only a small step.

\end{document}